\PassOptionsToPackage{table}{xcolor}
\documentclass[11pt]{article}

\usepackage{microsoft-tech-report}
\usepackage{microtype}
\usepackage{graphicx}
\usepackage{subcaption}
\usepackage{booktabs}
\usepackage{xcolor}
\usepackage{hyperref}
\usepackage{url}
\usepackage[utf8]{inputenc}
\usepackage[T1]{fontenc}
\usepackage{amsmath}
\usepackage{amssymb}
\usepackage{amsfonts}
\usepackage{mathtools}
\usepackage{amsthm}
\usepackage{bm}
\usepackage{nicefrac}
\usepackage{enumitem}
\usepackage{multirow}
\usepackage[capitalize,noabbrev]{cleveref}
\usepackage{xspace}
\usepackage{pifont}
\usepackage{wrapfig}
\usepackage{algorithm}
\usepackage{algpseudocode}
\algnewcommand{\algorithmicinput}{\textbf{Input:}}
\algnewcommand{\algorithmicoutput}{\textbf{Output:}}
\algnewcommand{\Input}{\item[\algorithmicinput]}
\algnewcommand{\Output}{\item[\algorithmicoutput]}
\usepackage{placeins}
\usepackage{tikz}
\usetikzlibrary{arrows.meta,positioning,shapes.geometric}
\usepackage[most]{tcolorbox}

\newcommand{\ourmethod}{SkillSpec\xspace}
\newcommand{\skillspecval}{SkillSpec-I\xspace}
\newcommand{\skillopt}{SkillOpt\xspace}
\newcommand{\skilloptlite}{SkillOpt-lite\xspace}
\newcommand{\na}{--}

\definecolor{ghaddbg}{HTML}{DAFBE1}
\definecolor{ghaddfg}{HTML}{116329}
\definecolor{ghdelbg}{HTML}{FFD7D5}
\definecolor{ghdelfg}{HTML}{82071E}
\definecolor{ghhunkbg}{HTML}{DDF4FF}
\definecolor{ghhunkfg}{HTML}{0969DA}
\definecolor{ghmetabg}{HTML}{F6F8FA}
\definecolor{ghmetafg}{HTML}{57606A}

\newcommand{\diffrow}[3]{%
  \noindent\colorbox{#1}{%
    \parbox{\dimexpr\linewidth-2\fboxsep\relax}{%
      \ttfamily\scriptsize\color{#2}\raggedright\strut #3\strut
    }%
  }\par
}
\newcommand{\DiffAdd}[1]{\diffrow{ghaddbg}{ghaddfg}{#1}}
\newcommand{\DiffDel}[1]{\diffrow{ghdelbg}{ghdelfg}{#1}}
\newcommand{\DiffHunk}[1]{\diffrow{ghhunkbg}{ghhunkfg}{#1}}
\newcommand{\DiffMeta}[1]{\diffrow{ghmetabg}{ghmetafg}{#1}}
\newcommand{\DiffContext}[1]{\diffrow{white}{black}{#1}}

\definecolor{deltapos}{HTML}{0E8A4A}
\definecolor{deltaneg}{HTML}{B91C1C}

\techreportlabel{Microsoft Research}
\techreportshorttitle{SkillSpec}

\hypersetup{
  colorlinks=true,
  linkcolor=msftblue,
  citecolor=msftblue,
  urlcolor=msftblue,
  pdfauthor={Huancheng Chen, Xiaodi Sun, Zhaoqiong Huang, Shenyang Huang, Shreya Singhal, Jingwen Lu},
  pdftitle={SkillSpec: Consensus-Gated Agent Skill Evolution via Representation Specialization}
}

\begin{document}
\thispagestyle{empty}

\noindent
\begin{minipage}[c]{0.5\linewidth}
\raggedright
\raisebox{-0.5\height}{\msftbrandmark}
\end{minipage}%
\begin{minipage}[c]{0.49\linewidth}
\raggedleft
{\msftdatefont\small\color{msftgray}October 2026}
\end{minipage}\par
\vspace{0.35em}
\noindent{\color{msftline}\rule{\linewidth}{0.8pt}\par}

\vspace{1.0em}
\begin{center}
{\msfttitlefont\fontsize{21}{25}\selectfont\color{msftdark}
SkillSpec: Consensus-Gated Agent Skill\\
Evolution via Representation Specialization\par}
\vspace{1.25em}

{\small\rmfamily\color{msftdark}
\mbox{Huancheng Chen \hspace{0.2em}
Xiaodi Sun \hspace{0.2em}
Zhaoqiong Huang \hspace{0.2em}
Shenyang Huang \hspace{0.2em}
Shreya Singhal \hspace{0.2em}
Jingwen Lu}\par}
\vspace{0.22cm}
{\footnotesize\rmfamily\color{msftgray}Microsoft AI\par}
\end{center}

\vspace{0.45em}
\begin{msfttitlebox}
\setlength{\parindent}{0cm}
\setlength{\parskip}{0.14cm}
\raggedright
\nohyphens

Natural-language \emph{skills} are textual procedural memories through which large
language model (LLM) agents retain reusable task knowledge without updating model
weights. Existing methods typically treat skills as either static artifacts or
monolithic documents optimized using aggregate validation scores as feedback. However,
representing a skill as a monolithic document restricts optimization to its
textual content, without explicitly modeling the structure through which
procedural knowledge is retrieved and executed. We identify a key distinction
between learning what knowledge to retain and determining how to organize it:
textual updates should first be validated through execution evidence, after
which the retained knowledge should be structured according to its procedural
dependencies and retrieval requirements. To this end, we introduce \textbf{SkillSpec},
a two-phase framework comprising consensus-gated evolution and representation
specialization. In the consensus-gated phase, complementary editing intents
generate complete candidate skills. An update is committed only when paired
evaluations reach consensus, requiring sufficient overall improvement and
non-negative aggregate paired gain in every repeated evaluation. In the
specialization phase, signals of process and redundancy sensitivity derived
from the full optimization trajectory, including accepted and rejected
candidates, guide the selection of a \emph{flat}, \emph{graph}, or
\emph{hybrid} representation. The validated backbone remains immutable, while
a derived artifact is initialized and specialized in the selected
representation. Across six benchmarks
and four target language models, SkillSpec improves average success rate over
SkillOpt by an absolute margin of \textbf{6.53\%}, averaged across the four models.
These results demonstrate that reliable skill evolution and representation
specialization address complementary objectives: deciding what knowledge to
retain and how to structure it for inference.

\vspace{0.14cm}
{\setlength{\parskip}{0.06cm}\small
{\msftmetalabel{Correspondence}
\href{mailto:huancchen@microsoft.com}{huancchen@microsoft.com}\par}
}
\end{msfttitlebox}

\section{Introduction}
\label{sec:introduction}

Large language model (LLM) agents increasingly use natural-language skills to
retain and reuse task knowledge without updating model weights. A \emph{skill} is a
textual procedural memory that encodes reusable steps, task rules, tool-use
guidance, answer formats, and error-recovery strategies while remaining easy
to inspect and revise.
However, a fixed skill cannot anticipate every task-specific failure:
execution trajectories may reveal missing procedures, conflicting guidance,
or underspecified constraints. Recent methods therefore treat prompts and
skills as optimizable state
\citep{zhou2022large,yang2023large,khattab2023dspy,
yuksekgonul2024textgrad,agrawal2025gepa,skillopt2026}, replacing one-shot
construction with iterative revision informed by execution feedback.

SkillOpt~\citep{skillopt2026} takes an important step toward controlled skill
evolution through bounded, validation-gated edits to a trainable skill
document. Nevertheless, aggregate validation scores can obscure an edit's
behavioral effects: improvements on some examples may conceal regressions on
others, and apparent gains may not persist across repeated evaluations.
Existing optimizers also preserve a single-document representation, changing
only its wording while leaving the organization of knowledge fixed. Flat
representations can obscure procedural dependencies in sequential tasks,
whereas unnecessary structure can introduce spurious dependencies and
irrelevant context. Our key observation is that deciding what knowledge to
retain and how to organize it are distinct optimization problems. The former
requires reliable evidence that an edit improves behavior, whereas the latter
requires evidence about procedural dependencies and interference. Addressing
both in a single rewriting loop makes it difficult to attribute gains or
choose an appropriate representation.

We introduce \textbf{SkillSpec}, a two-phase framework that separates reliable content
evolution from representation specialization. In the first phase, SkillSpec
generates candidate skills under four complementary editing intents:
\emph{repair}, \emph{preserve}, \emph{simplify}, and \emph{rewrite}. It
compares each candidate with the current skill on the same validation
instances. The consensus gate accepts an update only when it
improves performance overall with non-negative aggregate paired gain in every
repeated evaluation, while
retaining evidence from rejected candidates to inform representation
specialization. Building on the validated skill and the trajectory of
candidate evaluations, the second phase estimates process and redundancy
sensitivity, then uses these signals to select a \emph{flat}, \emph{graph}, or
\emph{hybrid} representation. Phase~II preserves the validated backbone as an
immutable source and specializes a derived artifact, allowing both its
guidance and organization to evolve within the selected representation.

We evaluate SkillSpec on six benchmarks, SearchQA~\citep{dunn2017searchqa},
DocVQA~\citep{mathew2021docvqa},
LiveMath~\citep{he2026livemathematicianbenchlivebenchmarkmathematicianlevel},
ALFWorld~\citep{alfworld}, SpreadsheetBench~\citep{spreadsheetbench}, and
OfficeQA~\citep{opsahl2026officeqa}, using four frontier language models.
Across the six benchmarks, SkillSpec improves average success rate over the
no-skill baseline by up to 27.53\%. Compared with
SkillOpt~\citep{skillopt2026}, SkillSpec yields gains of 12.26\%, 6.59\%, 3.12\%, and
4.14\% for GPT--4.1, GPT--5.4 Nano, GPT--5.4, and GPT--5.6 Sol,
respectively, averaging 6.53\% across the four models. These
results show the largest gain on GPT--4.1, suggesting that explicit skill
evolution and reorganization can remain valuable even when the underlying
model is less capable.
Our contributions can be summarized as follows:
\begin{itemize}[leftmargin=1.4em]
    \item We introduce consensus-gated skill evolution, which combines
    multi-intent candidate generation with paired validation to accept reliable
    content updates, reject edits that fail the aggregate gain thresholds, and
    retain evidence from rejected
    candidates to inform representation specialization.
    \item We introduce trajectory-guided specialization, which uses Phase~I
    evidence to select and optimize a \emph{flat}, \emph{graph}, or
    \emph{hybrid} representation while preserving the validated backbone as an
    immutable source.
    \item We evaluate SkillSpec across six benchmarks and four frontier
    language models, outperforming SkillOpt~\citep{skillopt2026} by 6.53\%
    on average across the four models.
\end{itemize}

\section{Related Work}
\label{sec:related_work}

\paragraph{Prompt optimization.}
A growing line of work treats natural-language prompts as optimizable artifacts
rather than static instructions. Search-based methods use task-level scores to
optimize candidate instructions~\citep{zhou2022large,yang2023large}, while
DSPy compiles prompts and demonstrations for compound LM
programs~\citep{khattab2023dspy}. TextGrad~\citep{yuksekgonul2024textgrad}
propagates natural-language feedback through a compound system, and
GEPA~\citep{agrawal2025gepa} uses reflective trajectory feedback while
retaining Pareto-efficient candidates over validation instances. These methods
establish execution feedback as a practical optimization signal, but operate
on prompts or LM programs rather than persistent agent skills.

\paragraph{Skill evolution.}
SkillOpt~\citep{skillopt2026} extends iterative optimization from prompts to
persistent skills, using execution trajectories to propose bounded revisions
and held-out evaluations to select updates.
SkillOpt-Lite~\citep{shen2026skilloptlite} replaces tree-based candidate
merging with direct revision of file-based trajectories while retaining an
independent validation gate. Subsequent methods prioritize candidates with
greater refinement potential~\citep{shastri2026bonsai}, stabilize update
directions and adapt edit budgets~\citep{li2026skilladam}, or train skill
editors directly from execution
feedback~\citep{yang2026skillmaster,vishe2026skillr1,peng2026writeexecuterefine}.
Other work repairs weak initial skills~\citep{liu2026skillrevise} or supports
long-horizon repository curation~\citep{ouyang2026skillos}. Methods in this
line improve how skill content is generated, selected, and maintained within a
predefined artifact schema. SkillSpec builds on feedback-driven skill
evolution with multi-intent candidates and paired, repeated validation, then
reuses outcomes from accepted and rejected candidates to inform representation
selection.

\paragraph{Skill representation.}
Prior work structures skills at different granularities.
Skill-Pro~\citep{procmem} represents an executable skill through initiation,
execution, and termination conditions. SkillX~\citep{wang2026skillx} builds a
multi-level skill hierarchy and iteratively refines its knowledge base, while
HyperSkill~\citep{xu2026hyperskill} organizes subtasks and reusable skills in
an evolving hypergraph for retrieval and maintenance.
Graph-of-Skills~\citep{liu2026graphofskills} instead constructs inter-skill
dependency graphs for inference-time structural retrieval, and
AIP~\citep{blumenfeld2026aip} compiles prose skills into schema-validated
execution graphs. Most closely, AutoRefine~\citep{autorefine} selects among
rules, skills, and bounded subagents by compiling trajectory-derived
interventions according to runtime ownership. Unlike methods that adopt a
predefined structure or select among artifact types, SkillSpec uses outcome
telemetry from accepted and rejected Phase~I candidates to select a
\emph{flat}, \emph{graph}, or \emph{hybrid} representation for a validated
skill before specialization.

\begin{figure*}[t]
    \centering
    \includegraphics[width=\textwidth]{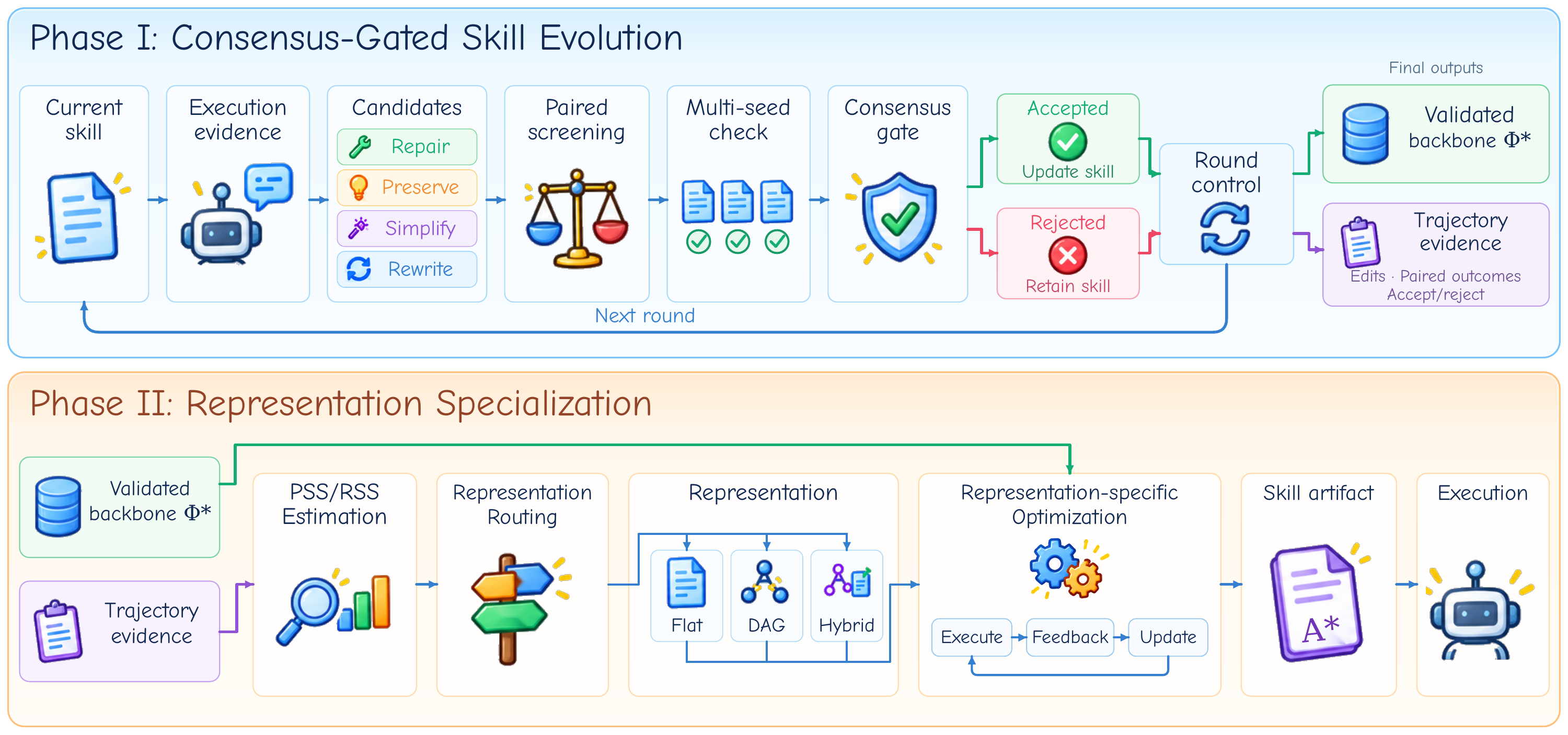}
    \caption{Overview of SkillSpec. Phase~I combines multi-intent candidate
    generation with repeated paired validation to produce a validated backbone
    and optimization trajectory. Phase~II estimates PSS and RSS from accepted
    and rejected candidates, selects a \emph{flat}, \emph{graph}, or
    \emph{hybrid} representation, and specializes a derived artifact while
    preserving the source backbone.}
    \label{fig:skillspec_a3_workflow}
\end{figure*}
\section{SkillSpec: A Two-Phase Framework for Skill Evolution}
\label{sec:method}

SkillSpec optimizes an external skill artifact \(\mathcal{A}\) for a fixed
policy model \(\pi_\theta\) without updating the model parameters \(\theta\).
For an instance \(x\), executing \(\pi_\theta\) conditioned on
\(\mathcal{A}\) produces a trajectory \(\tau_x(\mathcal{A})\) and evaluator
score \(s(\pi_\theta,x,\mathcal{A})\). Training trajectories provide evidence
for candidate revisions, while held-out validation outcomes determine which
revisions are retained, with the goal of improving performance on unseen test
instances.

Given an initial textual artifact \(\Phi_0\), SkillSpec decomposes artifact
optimization into content evolution and representation specialization. Phase~I
uses training and validation evidence to produce a validated textual backbone
\(\Phi^\star\) and an optimization trajectory \(\Gamma_{\mathrm{I}}\)
containing candidate artifacts, execution evidence, paired gains, and
validation decisions.
Phase~II estimates structural sensitivity from \(\Gamma_{\mathrm{I}}\) and selects a
representation
\(r^\star\in\{\emph{flat},\emph{graph},\emph{hybrid}\}\). It then initializes a
representation-specific artifact from \(\Phi^\star\) and specializes it to
obtain \(\mathcal{A}^\star\).
\Cref{fig:skillspec_a3_workflow} illustrates the two-phase workflow and the
artifacts passed from Phase~I to Phase~II.

\subsection{Phase I: Consensus-Gated Skill Evolution}
\label{sec:a21}

Phase~I evolves \(\Phi_0\) over \(T\) optimization rounds. At each round, the
policy model executes the current skill on a training batch. Successful and
failed trajectories then guide candidate generation under complementary
editing intents. Candidates are screened against the current skill using
task-level paired gains, and only updates confirmed across multiple executions
are committed. The final skill is returned as the validated backbone
\(\Phi^\star\).

\paragraph{Collecting execution evidence.}
At round \(t\), the fixed policy model uses the current skill \(\Phi_t\) to
solve a sampled training batch
\(\mathcal{B}_t\subset\mathcal{D}_{\mathrm{tr}}\). Phase~I partitions the
resulting trajectories into successful and failed executions according to
their evaluator outcomes. These trajectories and the associated evaluator
feedback form the execution evidence \(\mathcal{E}_t\) used to propose
candidate updates.

\paragraph{Multi-intent candidate generation.}
Given \(\mathcal{E}_t\), the skill optimizer constructs candidate skills under
the intent set
\(\mathcal{U}=\{\emph{repair},\emph{preserve},\emph{simplify},\emph{rewrite}\}\).
\emph{Repair} adds corrective guidance for recurring failures;
\emph{preserve} retains guidance supported by successful trajectories;
\emph{simplify} removes redundant or conflicting guidance; and \emph{rewrite}
restructures the skill when local edits are insufficient. Each intent yields a
complete skill candidate rather than an isolated edit. Compatible changes from
different intents can also be combined into an additional candidate. The
resulting candidate pool is
\(\mathcal{S}_t=\{\widetilde{\phi}_{t,j}\}_j\), where \(j\) indexes the
candidates generated at round \(t\). Keeping these candidates distinct allows
Phase~I to evaluate each update strategy independently rather than fold all
proposed changes into a single rewrite.

\paragraph{Screening with paired outcomes.}
Each candidate \(\widetilde{\phi}_{t,j}\in\mathcal{S}_t\) and the current skill
\(\Phi_t\) are evaluated on the same validation instances. Their evaluator
scores are compared for each \(x_i\in\mathcal{D}_{\mathrm{val}}\), yielding
the paired counts
\begin{equation}
\begin{aligned}
    n^+_{t,j}
        &= \sum_{x_i\in\mathcal{D}_{\mathrm{val}}}
           \mathbf{1}\!\left[
           s(\pi_\theta,x_i,\widetilde{\phi}_{t,j})
           >s(\pi_\theta,x_i,\Phi_t)
           \right],\\
    n^-_{t,j}
        &= \sum_{x_i\in\mathcal{D}_{\mathrm{val}}}
           \mathbf{1}\!\left[
           s(\pi_\theta,x_i,\widetilde{\phi}_{t,j})
           <s(\pi_\theta,x_i,\Phi_t)
           \right].
\end{aligned}
\label{eq:paired-screening}
\end{equation}
Here, \(n^+_{t,j}\) and \(n^-_{t,j}\) count task-level improvements and
regressions, respectively; ties contribute to neither count. Their difference
\(g_{t,j}=n^+_{t,j}-n^-_{t,j}\) is the net paired gain. Retaining the paired
improvement and regression counts exposes instance-level trade-offs, even
though their net difference is proportional to the accuracy difference under
binary scoring. Phase~I ranks candidates in descending order of \(g_{t,j}\), resolves
equal gains deterministically, and sends the highest-ranked \(N_f\) candidates
to the confirmation stage.

\paragraph{Confirming and committing an update.}
For each finalist \(\widetilde{\phi}_{t,j}\), let \(g_{t,j}^{(k)}\) denote its
paired gain in evaluation \(k\in\{0,\ldots,K\}\) on the complete
\(\mathcal{D}_{\mathrm{val}}\). We set \(g_{t,j}^{(0)}=g_{t,j}\) to reuse the
screening result, while \(k=1,\ldots,K\) index additional confirmation
executions. Let \(G_{\mathrm{pool}}\) denote the required pooled gain and
\(G_{\min}\) the minimum gain permitted in any single execution. Candidate
\(\widetilde{\phi}_{t,j}\) is confirmed only if
\begin{equation}
    \sum_{k=0}^{K}g_{t,j}^{(k)} \geq G_{\mathrm{pool}}
    \qquad\text{and}\qquad
    \min_{0\leq k\leq K} g_{t,j}^{(k)} \geq G_{\min}.
    \label{eq:a21-gate}
\end{equation}
The pooled-gain condition requires sufficient overall improvement, while the
minimum-gain condition ensures that every execution meets \(G_{\min}\).
With \(G_{\min}=0\), this requires non-negative aggregate paired gain in
each execution; individual validation instances may still regress.
If
multiple candidates satisfy both conditions, Phase~I selects the candidate with
the largest pooled gain, resolves ties deterministically, and denotes it by
\(\widetilde{\phi}_t^\star\). The update is then applied atomically:
\begin{equation}
\Phi_{t+1} =
\begin{cases}
\widetilde{\phi}_t^\star,
    & \text{if } \widetilde{\phi}_t^\star
      \text{ passes \eqref{eq:a21-gate}},\\
\Phi_t, & \text{otherwise}.
\end{cases}
\label{eq:atomic-update}
\end{equation}
Let \(\Omega_t\) denote the optimizer state used for candidate generation at
round \(t\). When a candidate is committed, Phase~I atomically updates both the
skill and this state; otherwise, it retains \((\Phi_t,\Omega_t)\). After each
round, it appends the candidate artifacts, execution evidence, paired gains,
and validation decisions to the optimization trajectory. After \(T\) rounds,
Phase~I returns the validated backbone \(\Phi^\star\) and complete trajectory
\(\Gamma_{\mathrm{I}}\).
\Cref{alg:phase1} summarizes the full update loop.

\begin{algorithm}[t]
\caption{Phase I: consensus-gated skill evolution}
\label{alg:phase1}
\begin{algorithmic}[1]
\small
\Input Fixed policy model \(\pi_\theta\), initial textual artifact \(\Phi_0\),
    training data \(\mathcal{D}_{\mathrm{tr}}\), validation data
    \(\mathcal{D}_{\mathrm{val}}\), round budget \(T\), finalist count \(N_f\),
    number \(K\) of confirmation executions, thresholds
    \(G_{\mathrm{pool}},G_{\min}\), and intent set \(\mathcal{U}\)
\Output Validated backbone \(\Phi^\star\) and optimization trajectory
    \(\Gamma_{\mathrm{I}}\)
\State Initialize optimizer state \(\Omega_0\) and trajectory
    \(\Gamma_0\gets\varnothing\)
\For{\(t=0,\ldots,T-1\)}
    \State Draw \(\mathcal{B}_t\subset\mathcal{D}_{\mathrm{tr}}\);
        \(\mathcal{E}_t\gets
        \Call{CollectEvidence}{\pi_\theta,\Phi_t,\mathcal{B}_t}\)
    \State \(\mathcal{S}_t\gets
        \Call{GenerateCandidates}{\Phi_t,\Omega_t,\mathcal{E}_t,\mathcal{U}}\),
        including compatible composition
    \ForAll{\(\widetilde{\phi}_{t,j}\in\mathcal{S}_t\)}
        \State Evaluate \(\widetilde{\phi}_{t,j}\) and \(\Phi_t\) on
            \(\mathcal{D}_{\mathrm{val}}\); compute \(g_{t,j}\) using
            \Cref{eq:paired-screening}
    \EndFor
    \State \(\mathcal{F}_t\gets\) top \(N_f\) candidates ranked by
        \(g_{t,j}\)
    \ForAll{\(\widetilde{\phi}_{t,j}\in\mathcal{F}_t\)}
        \State Set \(g_{t,j}^{(0)}\gets g_{t,j}\); run \(K\) additional paired
            evaluations on \(\mathcal{D}_{\mathrm{val}}\) to obtain
            \(\{g_{t,j}^{(k)}\}_{k=1}^{K}\)
    \EndFor
    \State \(\mathcal{Q}_t\gets
        \{\widetilde{\phi}_{t,j}\in\mathcal{F}_t:
        \sum_{k=0}^{K}g_{t,j}^{(k)}\geq G_{\mathrm{pool}}
        \land \min_{0\leq k\leq K}g_{t,j}^{(k)}\geq G_{\min}\}\)
    \If{\(\mathcal{Q}_t\neq\varnothing\)}
        \State \(\widetilde{\phi}_t^\star\gets
            \arg\max_{\widetilde{\phi}_{t,j}\in\mathcal{Q}_t}
            \sum_{k=0}^{K}g_{t,j}^{(k)}\), breaking ties deterministically
        \State Atomically set
            \(\Phi_{t+1}\gets\widetilde{\phi}_t^\star\) and update
            \(\Omega_{t+1}\)
    \Else
        \State \((\Phi_{t+1},\Omega_{t+1})\gets(\Phi_t,\Omega_t)\)
    \EndIf
    \State Append candidate artifacts, execution evidence, paired gains, and
        validation decisions to \(\Gamma_t\), forming \(\Gamma_{t+1}\)
\EndFor
\State \Return \(\Phi^\star\gets\Phi_T\) and
    \(\Gamma_{\mathrm{I}}\gets\Gamma_T\)
\end{algorithmic}
\end{algorithm}

\subsection{Phase II: Telemetry-Guided Structural Specialization}
\label{sec:a3-specialization}

Phase~II takes the validated backbone \(\Phi^\star\) and trajectory
\(\Gamma_{\mathrm{I}}\) from Phase~I. Using \(\Gamma_{\mathrm{I}}\), it
estimates structural sensitivity, selects a representation, and specializes a
derived artifact initialized from \(\Phi^\star\) while keeping the Phase~I
outputs unchanged.

\begin{figure*}[t]
\centering
\includegraphics[width=\textwidth]{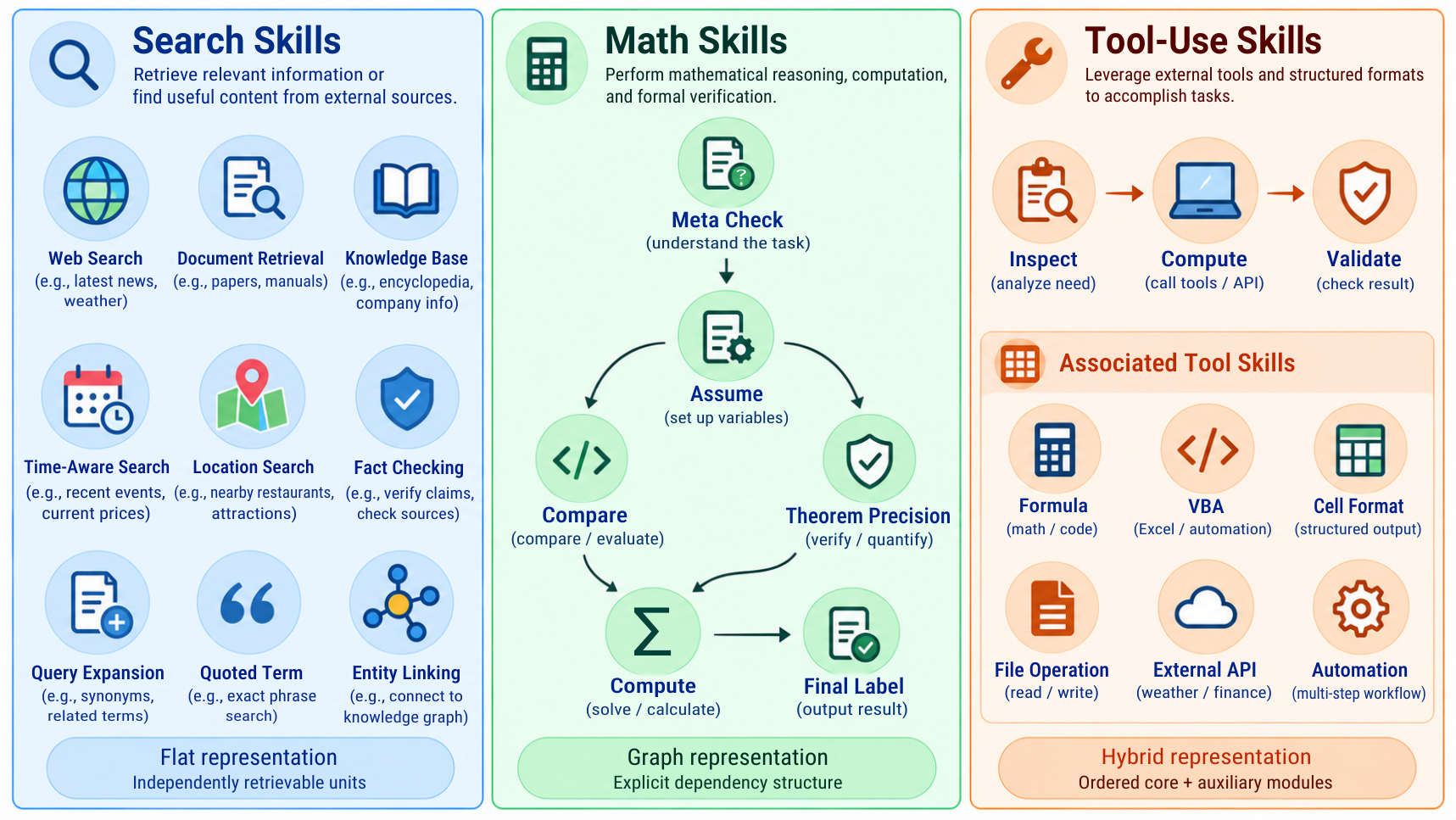}
\caption{Illustrative skill representations. \emph{Flat} organizes guidance
as independently retrievable units; \emph{graph} encodes explicit dependencies
among units; and \emph{hybrid} combines an ordered core with auxiliary modules.}
\label{fig:representation-examples}
\end{figure*}

\paragraph{Estimating structural sensitivity from telemetry.}
Let \(\mathcal{I}_{\mathrm{I}}\) index the candidates in
\(\Gamma_{\mathrm{I}}\) that differ from their round incumbent, pass patch
validation, and have a recorded screening gain. We refer to these candidates
as eligible and let \(\nu=|\mathcal{I}_{\mathrm{I}}|\). Phase~II summarizes
their structural sensitivity as
\begin{equation}
    \mathcal{T}=(\psi,\rho),
\label{eq:telemetry-output}
\end{equation}
where \(\mathcal{T}\) is a trajectory-derived structural profile used to
specialize the validated backbone, \(\psi\) is \emph{process sensitivity},
and \(\rho\) is \emph{redundancy sensitivity}.

Tasks differ in how their skill guidance should be organized. For example,
SearchQA~\citep{dunn2017searchqa} relies on independently retrievable
evidence-selection and answer-extraction guidance, whereas
LiveMath~\citep{he2026livemathematicianbenchlivebenchmarkmathematicianlevel}
benefits from an ordered procedure for formalizing and comparing theorem-style
claims. A fixed representation may therefore fail to capture necessary
dependencies or impose structure on otherwise independent guidance. SkillSpec
captures this trade-off with two signals:
\emph{process sensitivity} measures dependence on ordered steps, while
\emph{redundancy sensitivity} measures vulnerability to duplicated,
conflicting, or excessive guidance. Together, these signals guide the choice
among \emph{flat}, \emph{graph}, and \emph{hybrid} representations.

\paragraph{Process and redundancy sensitivity estimation.}
Each candidate \(\widetilde{\phi}_{t,j}\), \((t,j)\in
\mathcal{I}_{\mathrm{I}}\), contributes one training tuple
\((g_{t,j},p_{t,j},r_{t,j},c_{t,j})\). The label \(g_{t,j}\) is its screening
gain. The \emph{process} score \(p_{t,j}\) measures whether the revision adds
an ordered procedure whose later steps depend on earlier results; the
\emph{interference} score \(r_{t,j}\) measures duplicated, conflicting,
irrelevant, or excessively verbose guidance. Both use a four-level rubric:
0 (absent), 1 (weak or local), 2 (clear), and 3 (dominant). The normalized
edit size is
\begin{equation}
c_{t,j} =
D_{\mathrm{line}}(\Phi_t,\widetilde{\phi}_{t,j})
\,/\,\max\{L(\Phi_t),L(\widetilde{\phi}_{t,j})\},
\label{eq:normalized-edit-size}
\end{equation}
where \(D_{\mathrm{line}}\) counts inserted, deleted, or replaced lines and
\(L(\cdot)\) denotes the number of lines.

For each benchmark, ridge regression fits a sensitivity function \(f\) to
these tuples. Let \(f_{t,j}(p,r)\) be the predicted gain for candidate
\((t,j)\) after setting its \emph{process} and \emph{interference} scores to
\((p,r)\) and leaving its remaining inputs unchanged. Comparing mean
predictions with the zero-score baseline gives
\begin{equation}
\begin{aligned}
\Delta_{\mathrm{P}} &=
\max_{p\in\{1,2,3\}}
\frac{1}{\nu}\sum_{(t,j)\in\mathcal{I}_{\mathrm{I}}}
f_{t,j}(p,0)
-\frac{1}{\nu}\sum_{(t,j)\in\mathcal{I}_{\mathrm{I}}}
f_{t,j}(0,0),\\
\Delta_{\mathrm{R}} &=
\frac{1}{\nu}\sum_{(t,j)\in\mathcal{I}_{\mathrm{I}}}
f_{t,j}(0,0)
-\min_{r\in\{1,2,3\}}
\frac{1}{\nu}\sum_{(t,j)\in\mathcal{I}_{\mathrm{I}}}
f_{t,j}(0,r).
\end{aligned}
\label{eq:trajectory-contrasts}
\end{equation}
\(\Delta_{\mathrm{P}}\) compares the best nonzero \emph{process} level with
the baseline, while \(\Delta_{\mathrm{R}}\) measures the largest predicted
loss across nonzero \emph{interference} levels.

We construct \(N\) round-bootstrap datasets by sampling complete optimization
rounds with replacement. PSS and RSS are the fractions with
\(\Delta_{\mathrm{P}}>0\) and \(\Delta_{\mathrm{R}}>0\), respectively:
\begin{equation}
\psi=\frac{1}{N}\sum_{n=1}^{N}
\mathbf{1}\{\Delta_{\mathrm{P}}^{(n)}>0\},
\qquad
\rho=\frac{1}{N}\sum_{n=1}^{N}
\mathbf{1}\{\Delta_{\mathrm{R}}^{(n)}>0\}.
\label{eq:trajectory-sensitivity}
\end{equation}
PSS (\(\psi\)) therefore measures evidence that explicit \emph{process}
structure improves candidate outcomes, while RSS (\(\rho\)) measures evidence
that suppressing \emph{interference} improves them.

\paragraph{Adaptive representation selection.}
Phase~II selects among three structural representations: \emph{(a) flat} forms
a skill bank of independently retrievable units without an imposed execution
order; \emph{(b) graph} connects guidance nodes through explicit dependencies
to encode the required execution order; and \emph{(c) hybrid} combines an
ordered core with independently retrievable auxiliary modules.
\Cref{fig:representation-examples} illustrates these three structures.

The selector first uses PSS (\(\psi\)) to determine whether explicit ordering
is warranted. For high-PSS profiles, RSS (\(\rho\)) then distinguishes
\emph{graph} from \emph{hybrid}. RSS measures outcome evidence for harm from
\emph{interference}, not the prevalence of redundant guidance itself. Low PSS
therefore selects \emph{flat} to avoid imposing unsupported order. High PSS
selects an ordered representation: low RSS selects \emph{graph} to encode
dependencies directly, whereas high RSS selects \emph{hybrid} to preserve an
ordered core while modularizing redundancy-sensitive guidance.
Because PSS and RSS are bootstrap proportions, we set
\(\zeta_\psi=\zeta_\rho=0.50\), treating a sensitivity as present when a
majority of bootstrap datasets support its positive contrast.
\Cref{tab:representation-policy} summarizes the resulting decision rule and
the permissible specialization updates for each representation.
\begin{table*}[t]
\centering
\caption{Adaptive representation selection from PSS (\(\psi\)) and RSS
(\(\rho\)), where \(\zeta_\psi\) and \(\zeta_\rho\) denote the corresponding
thresholds and \(\Phi^\star\) denotes the frozen validated backbone.}
\label{tab:representation-policy}
\small
\renewcommand{\arraystretch}{1.22}
\begin{tabular}{@{}p{0.12\textwidth}@{\hspace{8pt}}p{0.19\textwidth}@{\hspace{8pt}}p{0.64\textwidth}@{}}
\toprule
\rowcolor{black!6}
\textbf{Structure} & \textbf{Condition} & \textbf{Specialization} \\
\midrule
\rowcolor{blue!5}
\emph{flat} & \(\psi\leq\zeta_\psi\) &
Initialize a derived copy from \(\Phi^\star\), split it into self-contained
guidance units, and add, revise, merge, or remove units without explicit
dependency edges or a prescribed order. \\
\rowcolor{green!5}
\emph{graph} & \(\psi>\zeta_\psi\), \(\rho\leq\zeta_\rho\) &
Initialize a derived graph from \(\Phi^\star\); jointly add, revise, or remove
nodes and directed dependency edges; and preserve a valid execution order. \\
\rowcolor{orange!6}
\emph{hybrid} & Otherwise &
Initialize an ordered core from \(\Phi^\star\); add, revise, or remove
auxiliary modules and update their attachment points. \\
\bottomrule
\end{tabular}
\end{table*}

\paragraph{Specialization procedure.}
Phase~II specializes the selected representation through iterative,
validation-gated updates. At each epoch, the fixed policy model \(\pi_\theta\)
executes the current artifact on a training batch and uses the resulting
trajectories and evaluator feedback to propose one revision, subject to the
representation-specific update rules in \Cref{tab:representation-policy}. The
revision may modify guidance, topology, retrieval metadata, or auxiliary
modules. The candidate is evaluated against the current artifact on the same
held-out examples and adopted only if it passes the Phase~II validation gate
detailed in Appendix~\ref{app:phase2_gate}; otherwise, the current artifact is
carried forward to the next epoch.

\providecommand{\ourmethod}{SkillSpec}
\providecommand{\skillspecval}{SkillSpec-I}
\providecommand{\skillopt}{\textsc{SkillOpt}}

\section{Experiments}
\label{sec:experiments}

\paragraph{Setup.}
Following the evaluation strategy of \skillopt{}~\citep{skillopt2026}, we
evaluate SkillSpec on GPT--4.1, GPT--5.4 Nano, GPT--5.4, and GPT--5.6 Sol across six
benchmarks: SearchQA~\citep{dunn2017searchqa},
SpreadsheetBench~\citep{spreadsheetbench},
OfficeQA~\citep{opsahl2026officeqa}, DocVQA~\citep{mathew2021docvqa},
LiveMath~\citep{he2026livemathematicianbenchlivebenchmarkmathematicianlevel},
and ALFWorld~\citep{alfworld}. We use each benchmark's training split for
optimization, its validation split to accept or reject candidate updates, and
its complete test split only for final evaluation. We report success rate per
benchmark and the macro average across all six benchmarks. Complete
hyperparameters and optimization budgets are reported in
\Cref{tab:a3_hyperparameters}.

\paragraph{Method comparison.}
We compare against six baselines.
\textbf{(a) Baseline} executes the evaluated model without an external skill.
\textbf{(b) GEPA}~\citep{agrawal2025gepa} reflects on execution trajectories
and retains Pareto-efficient prompt candidates.
\textbf{(c) Trace2Skill}~\citep{ni2026trace2skill} distills trajectory-local
lessons into reusable skills. \textbf{(d) SkillAdam}~\citep{li2026skilladam}
uses optimization memory and adaptive update magnitudes.
\textbf{(e) \skillopt{}}
~\citep{skillopt2026} iteratively optimizes a persistent skill.
\textbf{(f) \skilloptlite{}}~\citep{shen2026skilloptlite} performs file-based
skill optimization with Claude Sonnet~5.
We additionally report the consensus-gated Phase~I output, denoted
\skillspecval{}, as the paired reference for the complete Phase~II stage.

\subsection{Main Results}
\label{sec:main_results}

\begin{table*}[t]
\centering
\caption{Success rate (\%) on the complete test set and six-benchmark macro averages.
Improvement is the absolute success-rate difference (\%) from the no-skill baseline.
\skillspecval{} denotes the consensus-gated Phase~I output, and \ourmethod{}
denotes its Phase~II specialized result.
The GPT--4.1 LiveMath pair uses the legacy confidence-gated routing protocol,
which selected Flat, with GPT--4.1 as the Phase~II optimizer.
Best and second-best results are bolded and underlined, respectively.}
\label{tab:main_results}
\scriptsize
\setlength{\tabcolsep}{3.0pt}
\resizebox{\textwidth}{!}{%
\begin{tabular}{@{}llcccccccc@{}}
\toprule
\textbf{Model} & \textbf{Method} & \textbf{SearchQA} &
\textbf{Spreadsheet} & \textbf{OfficeQA} & \textbf{DocVQA} &
\textbf{LiveMath} & \textbf{ALFWorld} & \textbf{Average} &
\textbf{Improvement} \\
\midrule
\multirow{8}{*}{GPT--4.1}
 & Baseline & 69.57 & 36.07 & 1.16 & 66.31 & 25.00 & 44.78 & 40.48 & \na \\
 & GEPA & 79.71 & 35.36 & 12.79 & 80.75 & 29.03 & 64.93 &
 50.43 & \textcolor{deltapos}{+9.95} \\
 & Trace2Skill & 75.50 & 41.79 & 13.95 & 78.88 & 33.06 & 44.78 &
 47.99 & \textcolor{deltapos}{+7.51} \\
 & SkillAdam & 80.93 & 46.79 & 10.47 & 82.62 & 29.03 & 55.22 &
 50.84 & \textcolor{deltapos}{+10.36} \\
 & \skillopt{} & 79.93 & 46.43 & 11.05 & 73.26 & 29.03 & 52.99 & 48.78 &
 \textcolor{deltapos}{+8.30} \\
 & \skilloptlite{} & 75.71 & \underline{48.93} & 12.21 & 69.79 & 28.23 & 64.93 & 49.96 &
 \textcolor{deltapos}{+9.48} \\
\cmidrule(lr){2-10}
 & \skillspecval{} & \underline{81.86} & 47.50 & \underline{15.70} &
 \underline{84.22} & \underline{33.87} & \underline{65.67} & \underline{54.80} &
 \textcolor{deltapos}{\underline{+14.32}} \\
\rowcolor{blue!8}
 & \textbf{\ourmethod{}} & \textbf{82.43} & \textbf{50.36} &
 \textbf{27.91} & \textbf{84.49} & \textbf{41.94} &
 \textbf{79.10} & \textbf{61.04} & \textcolor{deltapos}{\textbf{+20.56}} \\
\midrule
\multirow{8}{*}{GPT--5.4 Nano}
 & Baseline & 56.36 & 30.71 & 1.74 & 29.14 & 27.42 & 28.36 & 28.95 & \na \\
 & GEPA & 67.36 & 30.71 & 5.81 & 64.44 & 26.61 & 39.55 &
 39.08 & \textcolor{deltapos}{+10.13} \\
 & Trace2Skill & 60.64 & 31.79 & 5.81 & 64.44 & 33.06 & 32.84 &
 38.10 & \textcolor{deltapos}{+9.15} \\
 & SkillAdam & 66.57 & 36.79 & 6.40 & 64.71 & 21.77 & 34.33 &
 38.43 & \textcolor{deltapos}{+9.48} \\
 & \skillopt{} & 57.43 & 39.64 & 5.81 & \underline{65.51} &
 \underline{50.81} & 33.58 & 42.13 & \textcolor{deltapos}{+13.18} \\
 & \skilloptlite{} & 64.00 & 41.43 & \underline{6.98} &
 \textbf{66.31} & 45.16 & \underline{44.78} & 44.78 & \textcolor{deltapos}{+15.83} \\
\cmidrule(lr){2-10}
 & \skillspecval{} & \underline{71.36} & \textbf{46.07} &
 \underline{6.98} & \textbf{66.31} & \underline{50.81} &
 \underline{44.78} & \underline{47.72} & \textcolor{deltapos}{\underline{+18.77}} \\
\rowcolor{blue!8}
 & \textbf{\ourmethod{}} & \textbf{71.79} & \underline{45.71} &
 \textbf{7.56} &
 \textbf{66.31} & \textbf{52.42} & \textbf{48.51} & \textbf{48.72} &
 \textcolor{deltapos}{\textbf{+19.77}} \\
\midrule
\multirow{8}{*}{GPT--5.4}
 & Baseline & 78.50 & 42.86 & 13.37 & 78.61 & 37.10 & 71.64 & 53.68 & \na \\
 & GEPA & 84.43 & 53.93 & \underline{65.12} & 80.21 & 42.74 & 84.33 &
 68.46 & \textcolor{deltapos}{+14.78} \\
 & Trace2Skill & 84.86 & 41.79 & 63.95 & \textbf{90.91} & 50.81 & 77.61 &
 68.32 & \textcolor{deltapos}{+14.64} \\
 & SkillAdam & 84.50 & 67.14 & 54.65 & \textbf{90.91} & 47.58 &
 83.58 & 71.39 & \textcolor{deltapos}{+17.71} \\
 & \skillopt{} & 84.00 & 71.43 & 59.88 & \underline{90.37} & 55.65 &
 \underline{89.55} & 75.15 & \textcolor{deltapos}{+21.47} \\
 & \skilloptlite{} & \underline{85.21} & \textbf{77.50} &
 63.37 & 89.84 & \underline{58.06} & 85.07 &
 \underline{76.51} & \textcolor{deltapos}{\underline{+22.83}} \\
\cmidrule(lr){2-10}
 & \skillspecval{} & 83.71 & 75.71 & 62.79 & 88.50 & 54.84 & 88.06 &
 75.60 & \textcolor{deltapos}{+21.92} \\
\rowcolor{blue!8}
 & \textbf{\ourmethod{}} & \textbf{86.36} & \underline{76.79} &
 \textbf{66.28} &
 88.77 & \textbf{58.87} & \textbf{92.54} & \textbf{78.27} &
 \textcolor{deltapos}{\textbf{+24.59}} \\
\midrule
\multirow{8}{*}{GPT--5.6 Sol}
 & Baseline & 73.07 & 36.79 & 18.60 & 63.90 & 50.00 & 85.82 & 54.70 & \na \\
 & GEPA & 79.71 & 37.14 & \textbf{68.02} & 89.04 & 65.32 & \underline{91.04} & 71.71 & \textcolor{deltapos}{+17.02} \\
 & Trace2Skill & 77.57 & 36.79 & 63.95 & \underline{91.71} & 62.10 & 85.82 & 69.66 & \textcolor{deltapos}{+14.96} \\
 & SkillAdam & 82.00 & 40.36 & 62.79 & 88.50 & 60.48 & \underline{91.04} & 70.86 & \textcolor{deltapos}{+16.17} \\
 & \skillopt{} & \underline{85.57} & \underline{82.14} & 54.65 & 90.37 & 67.74 & 88.06 & 78.09 & \textcolor{deltapos}{+23.39} \\
 & \skilloptlite{} & 85.00 & 79.29 & 63.95 & 88.77 & \textbf{78.23} & \underline{91.04} & \underline{81.05} & \textcolor{deltapos}{\underline{+26.35}} \\
\cmidrule(lr){2-10}
 & \skillspecval{} & 85.29 & 79.64 & 59.30 & 88.24 & 71.77 & \textbf{94.03} & 79.71 & \textcolor{deltapos}{+25.01} \\
\rowcolor{blue!8}
 & \textbf{\ourmethod{}} & \textbf{86.29} & \textbf{82.50} & \underline{65.70} & \textbf{91.98} & \underline{76.61} & 90.30 & \textbf{82.23} & \textcolor{deltapos}{\textbf{+27.53}} \\
\bottomrule
\end{tabular}
}
\end{table*}

\paragraph{Performance across models.}
SkillSpec achieves the highest average success rate for all four evaluated
models: 61.04\% on GPT--4.1, 48.72\% on GPT--5.4 Nano, 78.27\% on
GPT--5.4, and 82.23\% on GPT--5.6 Sol. These scores improve over the corresponding no-skill baselines by absolute margins of
20.56\%, 19.77\%, 24.59\%, and 27.53\%, respectively. They also exceed the strongest
external baseline for each model by 10.20\% on GPT--4.1
(SkillAdam), 3.94\% on GPT--5.4 Nano (\skilloptlite{}), 1.76\% on
GPT--5.4 (\skilloptlite{}), and 1.18\% on GPT--5.6 Sol (\skilloptlite{}).
At the benchmark level, SkillSpec leads on all six benchmarks with GPT--4.1;
is best or tied-best on SearchQA, OfficeQA, DocVQA, LiveMath, and ALFWorld
with GPT--5.4 Nano; leads on SearchQA, OfficeQA, LiveMath, and ALFWorld with
GPT--5.4, where it also ranks second on SpreadsheetBench; and leads on
SearchQA, SpreadsheetBench, and DocVQA with GPT--5.6 Sol.
Even before specialization, \skillspecval{}
achieves a higher overall average than any external method
when averaged across all four models.
This result indicates that multi-intent optimization in Phase~I already
provides a strong foundation, while Phase~II yields additional gains on
average through representation-specific continued optimization.

\paragraph{Paired specialization gain.}
Across the 24 paired model--benchmark settings, SkillSpec outperforms its exact
\skillspecval{} backbone in 21, ties in one, and underperforms in two.
Using the reported macro averages, Phase~II yields absolute gains of \(+6.24\%\),
\(+1.00\%\), \(+2.67\%\), and \(+2.52\%\) for GPT--4.1, GPT--5.4 Nano,
GPT--5.4, and GPT--5.6 Sol, respectively. Since
each pair shares the same starting checkpoint and provenance, these results
capture the overall effect of Phase~II, including both representation
selection and continued optimization.
For the updated GPT--5.4 Nano SearchQA pair, hard accuracy increases from
999/1,400 (71.36\%) to 1,005/1,400 (71.79\%). Phase~II corrects 58 previously
incorrect answers and loses 52 previously correct answers, a net gain of six
answers (\(+0.43\) percentage points). The frozen Phase~I backbone and
Phase~II result are evaluated on identical test IDs, with no execution
failures or recorded timeouts; this is a single-run comparison.
For the updated GPT--4.1 DocVQA pair, Phase~II corrects 14 previously
incorrect answers but loses 13 previously correct answers, yielding a net
gain of one answer (\(+0.27\) percentage points) on 374 test items.
Both evaluations use the same test identities and have no agent failures
or recorded timeout failure reasons. This single-run result, with unpinned
upstream SkillOpt, does not establish a robust improvement; comparison
with older Sonnet-edited, capped runs also changes both the editor and
the delta-budget policy.

\subsection{Phase I Dynamics}
\label{sec:phase1_dynamics}

\begin{figure*}[t]
\centering
\includegraphics[width=\textwidth]{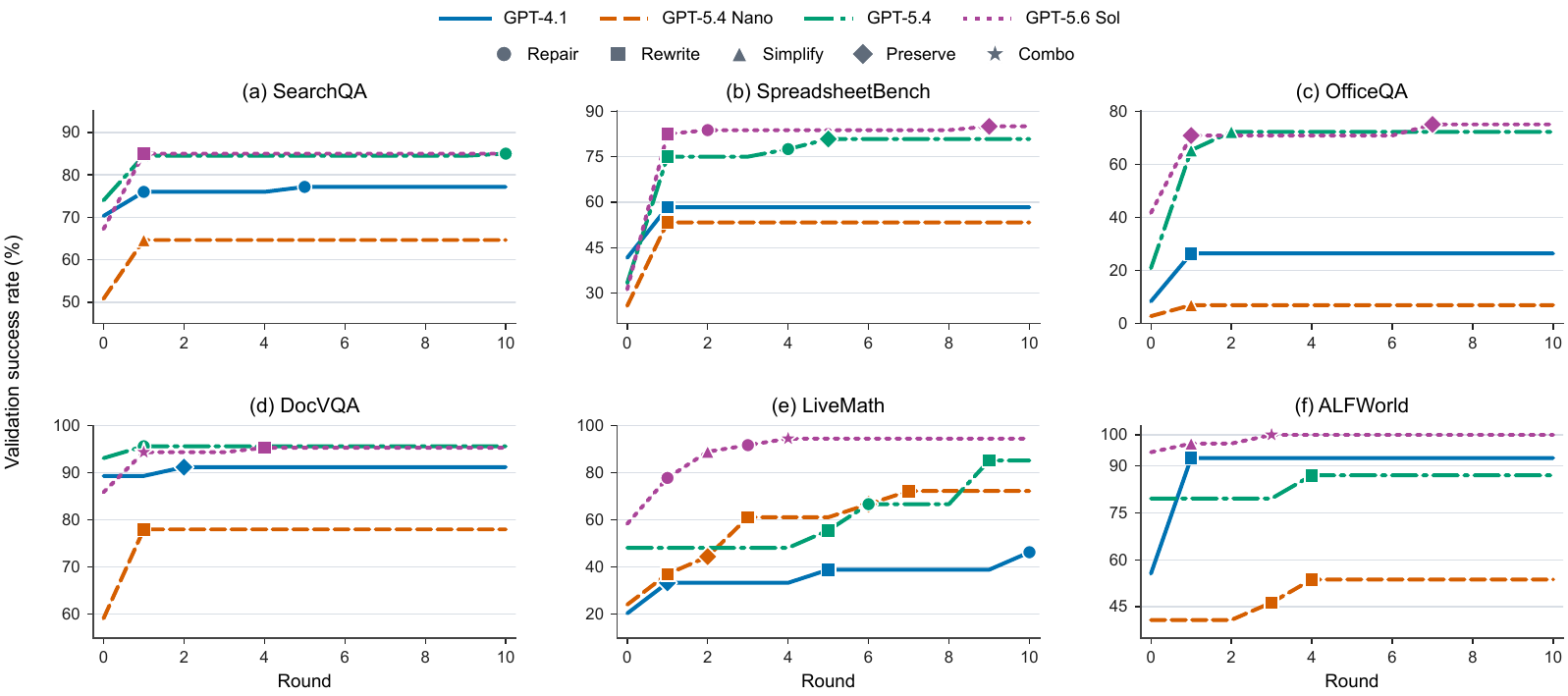}
\caption{Optimization trajectories during Phase~I across six benchmarks for
GPT--4.1, GPT--5.4 Nano, GPT--5.4, and GPT--5.6 Sol. All trajectories use
10 rounds, with three validation repeats for the first three models and two
for GPT--5.6 Sol.}
\label{fig:phase1_dynamics}
\end{figure*}

\Cref{fig:phase1_dynamics} suggests that accepted edits shift from behavioral
contract formation to local constraint refinement as the incumbent matures.
Early trajectories are shaped primarily by \emph{rewrite}, which accounts for
15 of 32 accepted updates across the three original models; these accepted updates
occur in five of the six benchmarks. The less frequent \emph{simplify} has
four accepted updates, all in rounds 1--2. Later gains increasingly come from
targeted changes: four of seven accepted \emph{repair} updates occur after
round~4, while \emph{preserve} appears mainly on LiveMath and the already
strong GPT--4.1 DocVQA trajectory.
Appendix~\ref{app:intent_rewrite}--\ref{app:intent_preserve} illustrates the
same shift: the early cases restructure workflows or output contracts, whereas
the later cases narrow or clarify specific exceptions.

\subsection{Representation Ablation}
\label{sec:ablations}

For GPT--5.4 Nano, \Cref{tab:ablations} reveals distinct Phase~I sensitivity
profiles across benchmarks.
\textbf{SearchQA} and \textbf{DocVQA} share a low-PSS, high-RSS profile,
indicating limited support for beneficial explicit ordering but evidence that
reducing redundant or conflicting guidance is beneficial.
\textbf{OfficeQA}, in contrast, involves retrieving relevant documents,
extracting evidence, and performing calculations that depend on the extracted
values. Unlike answering from supplied snippets in SearchQA or a document
image in DocVQA, this workflow introduces dependencies between intermediate
steps, consistent with its higher PSS and the use of Graph to encode their
order.
\textbf{SpreadsheetBench}, \textbf{LiveMath}, and \textbf{ALFWorld} share a
high-PSS, high-RSS profile in the Phase~I estimates reported in
\Cref{tab:ablations}, motivating both process structure and interference
control. Dependencies between spreadsheet operations, mathematical reasoning
steps, or state-dependent actions in ALFWorld motivate an ordered core,
while redundancy sensitivity motivates keeping auxiliary guidance modular.

\begin{table*}[t]
\centering
\caption{Controlled representation ablation on GPT--5.4 Nano, reported
separately from the main-table runs. Within each benchmark, Flat, Graph, and
Hybrid are initialized from the same Phase~I backbone and use the same
specialization budget. Boxes indicate the branches selected by PSS/RSS.
Avg.\ is the unweighted mean across six benchmarks.}
\label{tab:ablations}
\small
\setlength{\tabcolsep}{3.5pt}
\resizebox{\textwidth}{!}{%
\begin{tabular}{@{}lccccccc@{}}
\toprule
\textbf{Variant} & \textbf{SearchQA} & \textbf{Spreadsheet} &
\textbf{OfficeQA} & \textbf{DocVQA} & \textbf{LiveMath} &
\textbf{ALFWorld} & \textbf{Avg.} \\
\midrule
Phase~I, equal compute & 66.57 & 46.79 & 4.07 & 65.24 & 45.97 & 51.49 & 46.69 \\
Phase~II, Flat only & \fbox{\strut 68.50} & 46.79 & 4.65 &
  \fbox{\strut 66.58} & 41.94 & 52.99 & 46.91 \\
Phase~II, Graph only & 67.86 & 43.57 & \fbox{\strut 6.98} &
  67.11 & 45.97 & 55.97 & 47.91 \\
Phase~II, Hybrid only & 66.93 & \fbox{\strut 49.29} & 5.81 & 62.30 &
  \fbox{\strut 53.23} & \fbox{\strut 57.01} & 49.10 \\
\midrule
PSS (\(\psi\)) & 0.428 & 0.925 & 0.619 & 0.411 & 0.998 & 0.778 & \na \\
RSS (\(\rho\)) & 0.918 & 0.837 & 0.499 & 0.653 & 1.000 & 0.889 & \na \\
\textbf{Router} & Flat & Hybrid & Graph & Flat & Hybrid & Hybrid & \na \\
\rowcolor{blue!8}
\ourmethod{} & 68.50 & 49.29 & 6.98 & 66.58 & 53.23 & 57.01 & 50.27 \\
\bottomrule
\end{tabular}
}
\end{table*}

The \emph{Phase~I, equal compute} control retains the textual representation
and continues skill optimization with the same number of epochs and training
examples as each Phase~II variant; this matches the continuation schedule but
does not establish equal token usage or end-to-end computation.
Using only Phase~I telemetry to choose among
the specialization branches, SkillSpec selects the best-performing
representation on five of the six benchmarks. On the
remaining benchmark, DocVQA, it selects Flat, which trails Graph by just
0.53\% (66.58\% versus 67.11\%). Across all six benchmarks,
SkillSpec's six-benchmark macro-average is 50.27\%, above fixed Flat, Graph,
and Hybrid by absolute margins of 3.36\%, 2.36\%, and 1.17\%, respectively.
All aggregates are computed from unrounded per-benchmark scores. The
six-benchmark comparison supports adapting the representation to Phase~I
sensitivity rather than imposing a fixed structure, without consulting test
outcomes during routing.

\FloatBarrier
\section{Conclusion}

We introduced SkillSpec, a two-phase framework for improving LLM agents
through skill optimization without updating model weights. Phase~I uses
consensus-gated evolution to refine reusable knowledge, while Phase~II uses
trajectory-derived sensitivity signals to select and optimize a flat, graph,
or hybrid organization. Across six benchmarks, SkillSpec
improves the average success rate over SkillOpt by absolute margins of
12.26\%, 6.59\%, 3.12\%, and 4.14\% on GPT--4.1, GPT--5.4 Nano, GPT--5.4,
and GPT--5.6 Sol, respectively. These findings suggest that skill optimization
can benefit from considering both what knowledge an agent retains and
how that knowledge is structured for execution.

\bibliographystyle{unsrtnat}
\bibliography{references}

\clearpage
\appendix
\section{Hyperparameters and Configuration}
\label{app:hyperparameters}

\begin{table}[h]
\centering
\caption{Experimental and specialization settings used for SkillSpec unless
noted otherwise.}
\label{tab:a3_hyperparameters}
\small
\begin{tabular}{@{}p{0.40\columnwidth}p{0.54\columnwidth}@{}}
\toprule
\textbf{Parameter} & \textbf{Value} \\
\midrule
Evaluated models & GPT--4.1 / GPT--5.4 Nano / GPT--5.4 / GPT--5.6 Sol \\
Optimization data & Official training and validation splits \\
Final evaluation & Complete official test split \\
Primary metric & Hard accuracy \\
Aggregation & Macro average across six benchmarks \\
Failure accounting & Agent, setup, and timeout failures count as incorrect \\
\midrule
Initial skill &
Official benchmark seed skill for each model--benchmark cell \\
Phase I rounds / batch size & 10 / 40 \\
Candidate profiles &
\emph{repair}, \emph{preserve}, \emph{simplify}, \emph{rewrite} \\
Combo candidate & Optional, from compatible candidates \\
Primary comparison & Paired on identical task IDs \\
Finalists \(N_f\) & 2 \\
Total Phase I paired evaluations &
\(K+1\), including screening \\
Minimum pooled net gain \(G_{\mathrm{pool}}\) & 3 tasks \\
Minimum per-evaluation gain \(G_{\min}\) & 0 tasks \\
Phase II epochs / batch size & 4 / 35 \\
PSS/RSS estimator & Candidate-level ridge, \(\ell_2=1\) \\
Estimator controls &
Round, candidate profile, change ratio, quadratic process/interference \\
Estimator eligibility &
Changed, valid candidates with a recorded screening gain \\
Annotation evidence & Candidate skill diffs \\
Annotation scale & Integer process/interference strengths in \(\{0,1,2,3\}\) \\
Round-cluster bootstrap & 2,000 samples, seed 42 \\
PSS threshold \(\zeta_\psi\) & 0.50 \\
RSS threshold \(\zeta_\rho\) & 0.50 \\
Flat policy & \(\psi\le\zeta_\psi\) \\
Graph policy & \(\psi>\zeta_\psi\) and \(\rho\leq\zeta_\rho\) \\
Hybrid policy & Otherwise \\
Phase I / telemetry optimizer & Claude Sonnet 5 via Copilot CLI \\
\bottomrule
\end{tabular}
\end{table}

Representation decisions use only sanitized Phase~I telemetry and the accepted
backbone. The telemetry optimizer extracts structured observations; fixed code
computes PSS, RSS, and support from them. Benchmark and task identifiers are
prohibited as explicit decision features, and no test-set labels or outcomes
are provided. This restriction removes direct identifiers but does not imply
that semantic task characteristics are absent from the skill or trajectory.

\subsection{Phase II Validation Gate}
\label{app:phase2_gate}

Each Phase~II candidate is evaluated once on a fixed held-out selection set
using the same task IDs and execution seed across epochs. The gate metric is
hard accuracy. Let \(a_e\) denote the candidate accuracy at epoch \(e\) and
\(a_{\mathrm{inc}}\) the cached incumbent accuracy, initialized by evaluating
the initial Phase~II artifact on the same selection set. The candidate is
adopted if and only if \(a_e>a_{\mathrm{inc}}\), in which case the cached score
is updated to \(a_e\). Ties and regressions retain the incumbent. This gate is
distinct from the repeated paired consensus gate used in Phase~I.

\section{Phase I Intent Examples}
\label{app:intent_examples}

Below, we present representative diffs for one accepted update from each
intent. Added and removed lines are highlighted in green and red,
respectively, while unchanged content is omitted. Because \emph{rewrite} can
reorganize most of a skill, we show its principal edits rather than the full
diff. To make the before--after contrast explicit, we render the append-only
\emph{preserve} clarification as a normalized replacement: the original rule
is shown in red and the accepted rule, including the added clarification, in
green. For display, Unicode punctuation is normalized to ASCII and
percentage-unit wording in the \emph{simplify} diff is rendered as \%.
No other wording is changed. Reported validation scores are mean hard
accuracies over three repeats.

\subsection{\emph{Rewrite}}
\label{app:intent_rewrite}
\noindent\textbf{GPT--5.4 SpreadsheetBench, round~1.}
  \begin{tcolorbox}[
    enhanced,
    breakable,
    colback=white,
    colframe=black!18,
    colbacktitle=black!10,
    coltitle=black,
    fonttitle=\bfseries\small,
    title={Representative Skill Diff},
    boxrule=0.45pt,
    arc=1.5pt,
    boxsep=0pt,
    left=0pt,
    right=0pt,
    top=0pt,
    bottom=0pt,
    before skip=6pt,
    after skip=10pt
  ]
  \DiffHunk{@@ Common Workflow / formula handling @@}
\DiffDel{-4. Confirm the target cells/range contain the expected values.}
\DiffAdd{+4. Confirm the target cells/range contain the expected values by re-opening the}
\DiffAdd{+   saved output file and printing cell.value for every requested cell.}
\DiffAdd{+   A blank/None result means the task is not done yet.}
\DiffAdd{+}
\DiffAdd{+\#\# Formulas vs. Computed Values}
\DiffAdd{+}
\DiffAdd{+openpyxl never evaluates formulas. If you set cell.value = "=SUM(A1:A10)",}
\DiffAdd{+it writes only the formula string; it does not compute or cache a result.}
\DiffAdd{+}
\DiffAdd{+- Treat "write a formula for X" as "make the correct result appear in the}
\DiffAdd{+  target cell(s)", not as "the cell must literally contain an = string".}
\DiffAdd{+- Reproduce the formula logic in Python and write the resulting literal value}
\DiffAdd{+  with the correct type into every target cell.}
\DiffAdd{+- Apply the operation to every row or column covered by the instruction.}
\DiffAdd{+- Only leave an actual formula string when only its text will be inspected and}
\DiffAdd{+  no computed value is required.}
\DiffAdd{+}
\DiffAdd{+\#\# Cell Content Discipline}
\DiffAdd{+}
\DiffAdd{+- A target cell must contain only the requested value, never explanatory prose.}
\DiffAdd{+- Strip surrounding labels unless the instruction asks to retain them.}
\DiffAdd{+- Reopen the output and verify that requested cells are not blank or None.}

  \end{tcolorbox}%

\noindent\textbf{Before:} The skill only required checking that target cells
contained the expected values.\\
\textbf{After:} The rewrite introduced an
explicit reopen-and-verify procedure, distinguished formulas from computed
values, and required literal, nonblank outputs in every requested cell.

\noindent\textbf{Validation:} \(33.3\%\rightarrow75.0\%\).

\subsection{\emph{Simplify}}
\label{app:intent_simplify}
\noindent\textbf{GPT--5.4 OfficeQA, round~2.}
  \begin{tcolorbox}[
    enhanced,
    breakable,
    colback=white,
    colframe=black!18,
    colbacktitle=black!10,
    coltitle=black,
    fonttitle=\bfseries\small,
    title={Skill Diff},
    boxrule=0.45pt,
    arc=1.5pt,
    boxsep=0pt,
    left=0pt,
    right=0pt,
    top=0pt,
    bottom=0pt,
    before skip=6pt,
    after skip=10pt
  ]
  \DiffMeta{--- Original Skill}
\DiffMeta{+++ Accepted Skill}
\DiffHunk{@@ -14,3 +14,4 @@}
\DiffContext{ - Return the final answer only after one last consistency check against the retrieved evidence.}
\DiffContext{ - Copy the final answer from a checked value, not from an unverified intermediate guess.}
\DiffDel{-- Inside the final answer tags, state only the bare value the question asks for (plus a unit/word only if the requested format explicitly calls for one) --- no restated reasoning, source citations, intermediate steps, or explanatory sentences.}
\DiffAdd{+- Inside the final answer tags, state only the bare value the question asks for --- no restated reasoning, source citations, intermediate steps, or explanatory sentences.}
\DiffAdd{+- Do not append a unit, label, or direction word (e.g., "\%," "inflow," "outflow") just because the question's narrative describes the metric that way; a phrase like "report in millions" or "in \%" describes the number's scale, not literal text to add. Include such a word only when the output-format instruction explicitly asks the answer string itself to contain it (e.g., a named date format).}

  \end{tcolorbox}%

\noindent\textbf{Before:} A single rule combined the bare-value requirement
with an underspecified exception for units or words.\\
\textbf{After:} The update
separated the two concerns and clarified that narrative scale cues are not
output tokens unless the requested format explicitly includes them.

\noindent\textbf{Validation:} \(65.3\%\rightarrow72.2\%\).

\subsection{\emph{Repair}}
\label{app:intent_repair}
\noindent\textbf{GPT--5.4 SpreadsheetBench, round~4.}
  \begin{tcolorbox}[
    enhanced,
    breakable,
    colback=white,
    colframe=black!18,
    colbacktitle=black!10,
    coltitle=black,
    fonttitle=\bfseries\small,
    title={Skill Diff},
    boxrule=0.45pt,
    arc=1.5pt,
    boxsep=0pt,
    left=0pt,
    right=0pt,
    top=0pt,
    bottom=0pt,
    before skip=6pt,
    after skip=10pt
  ]
  \DiffMeta{--- Original Skill}
\DiffMeta{+++ Accepted Skill}
\DiffHunk{@@ -40,9 +40,13 @@}
\DiffContext{   formula the user manually copies down row-by-row must be replicated in full for all rows}
\DiffContext{   it applies to; don't stop after solving the first cell.}
\DiffDel{-- Only leave an actual `=` formula string in a cell when the instruction is explicitly about}
\DiffDel{-  formula authoring/repair *and* nothing will need to read that cell's computed value (e.g.}
\DiffDel{-  purely cosmetic/example formulas). When in doubt, prefer writing the literal computed}
\DiffDel{-  result --- it satisfies both "the formula works" and "the value is correct", while a bare}
\DiffDel{-  formula string alone satisfies neither once opened by a non-recalculating reader.}
\DiffAdd{+- Requests phrased as "I need a formula for X", "write a formula that...", or "how do I fix/}
\DiffAdd{+  adjust this formula" describe the desired *result*, not a literal instruction to store an}
\DiffAdd{+  `=` string --- treat them the same as any other request and write the literal computed value.}
\DiffAdd{+  Do **not** leave a bare `=` formula string in a target cell just because the instruction}
\DiffAdd{+  uses the word "formula": a reader that never runs a spreadsheet engine cannot distinguish}
\DiffAdd{+  an uncomputed formula cell from a blank one, so a bare formula string reads back as empty}
\DiffAdd{+  and fails verification exactly like leaving the cell untouched. Only keep a literal `=`}
\DiffAdd{+  string when the instruction asks you to inspect/repair the formula text itself (e.g. "what's}
\DiffAdd{+  wrong with this formula") and no cell's resulting value is referenced anywhere in the task.}
\DiffContext{ - Double-check edge cases the instruction calls out explicitly (blanks, `\#N/A`, zero values,}
\DiffContext{   duplicates, date/leap-year boundaries) --- these are frequently the exact cells checked.}

  \end{tcolorbox}%

\noindent\textbf{Before:} The skill broadly allowed literal formula strings
for formula-authoring or repair requests.\\
\textbf{After:} The repair narrowed
that exception to tasks that inspect formula text and otherwise required the
literal computed value.

\noindent\textbf{Validation:} \(75.0\%\rightarrow77.5\%\).

\subsection{\emph{Preserve}}
\label{app:intent_preserve}
\noindent\textbf{GPT--5.4 SpreadsheetBench, round~5.}
  \begin{tcolorbox}[
    enhanced,
    breakable,
    colback=white,
    colframe=black!18,
    colbacktitle=black!10,
    coltitle=black,
    fonttitle=\bfseries\small,
    title={Skill Diff},
    boxrule=0.45pt,
    arc=1.5pt,
    boxsep=0pt,
    left=0pt,
    right=0pt,
    top=0pt,
    bottom=0pt,
    before skip=6pt,
    after skip=10pt
  ]
  \DiffMeta{--- Original Skill}
\DiffMeta{+++ Accepted Skill}
\DiffHunk{@@ Formula-text exception @@}
\DiffDel{-- Requests phrased as "I need a formula for X", "write a formula that...", or "how do I fix/}
\DiffDel{-  adjust this formula" describe the desired *result*, not a literal instruction to store an}
\DiffDel{-  `=` string --- treat them the same as any other request and write the literal computed value.}
\DiffDel{-  Do **not** leave a bare `=` formula string in a target cell just because the instruction}
\DiffDel{-  uses the word "formula": a reader that never runs a spreadsheet engine cannot distinguish}
\DiffDel{-  an uncomputed formula cell from a blank one, so a bare formula string reads back as empty}
\DiffDel{-  and fails verification exactly like leaving the cell untouched. Only keep a literal `=`}
\DiffDel{-  string when the instruction asks you to inspect/repair the formula text itself (e.g. "what's}
\DiffDel{-  wrong with this formula") and no cell's resulting value is referenced anywhere in the task.}
\DiffAdd{+- Requests phrased as "I need a formula for X", "write a formula that...", or "how do I fix/}
\DiffAdd{+  adjust this formula" describe the desired *result*, not a literal instruction to store an}
\DiffAdd{+  `=` string --- treat them the same as any other request and write the literal computed value.}
\DiffAdd{+  Do **not** leave a bare `=` formula string in a target cell just because the instruction}
\DiffAdd{+  uses the word "formula": a reader that never runs a spreadsheet engine cannot distinguish}
\DiffAdd{+  an uncomputed formula cell from a blank one, so a bare formula string reads back as empty}
\DiffAdd{+  and fails verification exactly like leaving the cell untouched. Only keep a literal `=`}
\DiffAdd{+  string when the instruction asks you to inspect/repair the formula text itself (e.g. "what's}
\DiffAdd{+  wrong with this formula") and no cell's resulting value is referenced anywhere in the task.}
\DiffAdd{+  If the instruction names or implies where the result belongs (e.g. "it should be placed}
\DiffAdd{+  into cell K6", "fill out row 11", a specific `Sheet!Cell` reference), that counts as}
\DiffAdd{+  referencing the cell's resulting value --- the debug-only exception above does **not** apply}
\DiffAdd{+  even when the wording also asks you to diagnose/explain the formula, so compute and write}
\DiffAdd{+  the literal value there. Never leave such a target cell as `None`/blank as a fallback.}

  \end{tcolorbox}%

\noindent\textbf{Before:} The formula-text exception did not explicitly cover
diagnostic requests that also named a destination cell.\\
\textbf{After:} The
preserved rule was augmented to treat any named destination as requiring its
computed value and to forbid a blank fallback.

\noindent\textbf{Validation:} \(77.5\%\rightarrow80.8\%\).

\section{Specialization Examples}
\label{app:skill_examples}

Table~\ref{tab:skill_examples} shows condensed examples of procedural content
that may appear in the learned Flat, Graph, or Hybrid specialization.
These are not benchmark-specific prompts written by hand, and they do not imply
test-time selection among representation families. \ourmethod{} selects one
family per model--benchmark cell before specialization training.

\begin{table*}[t]
\centering
\caption{Representative specialization content. Text is lightly abbreviated for readability.}
\label{tab:skill_examples}
\small
\setlength{\tabcolsep}{4pt}
\begin{tabular}{@{}
    >{\raggedright\arraybackslash}p{0.16\linewidth}
    >{\raggedright\arraybackslash}p{0.18\linewidth}
    >{\raggedright\arraybackslash}p{0.25\linewidth}
    >{\raggedright\arraybackslash}p{0.33\linewidth}
    @{}}
\toprule
\textbf{Benchmark} & \textbf{Content role} & \textbf{Purpose} & \textbf{Example content} \\
\midrule
LiveMath & Procedural specialization & Distinguish theorem-style statements with subtle strength differences. & \emph{Before choosing, rewrite each option as a formal claim: hypotheses, quantifiers, conclusion strength, and endpoint cases. Rank options by logical implication, not by thematic similarity. Do not stop at the first true-looking statement; choose the strongest statement actually proved.} \\
\addlinespace
DocVQA & Evidence specialization & Preserve literal values and provenance from a document image. & \emph{Ground the answer in visible document evidence. Prefer the exact string, unit, date, or entity shown in the source. If multiple nearby values appear, compare labels and table headers before answering.} \\
\addlinespace
OfficeQA & Backbone-anchored process & Support multi-turn file/search tasks that require verified evidence. & \emph{Search broadly, then narrow by entity/date/table. Read the source document before answering. If evidence is missing, reformulate the query. Track provenance and answer only from retrieved evidence; do not answer from memory.} \\
\addlinespace
Spreadsheet\allowbreak Bench & Transformation specialization & Preserve unrelated workbook state during a requested edit. & \emph{Use the full workbook, not only the preview. Iterate over all actual rows and sheets. Preserve unrelated cells and formulas. Write outputs to the requested sheet/cell range and validate against the requested answer position.} \\
\addlinespace
ALFWorld & Backbone-anchored process & Preserve state across long action sequences and recovery steps. & \emph{Track the goal, current room, inventory, and object state. Explore systematically; after a failed action, inspect the observation and choose a recovery action. Use containers and object affordances explicitly; stop only after the goal is satisfied.} \\
\bottomrule
\end{tabular}
\end{table*}

\section{PSS/RSS Estimation}
\label{app:sensitivity_details}

This section specifies the candidate-level annotation and fixed
ridge-estimation procedure used for the sensitivity scores in
\Cref{tab:ablations}.
It complements \Cref{eq:trajectory-contrasts,eq:trajectory-sensitivity}:
the language model supplies evidence-grounded labels, while numerical code
fits the outcome model and computes the scores.

\subsection{Annotation evidence and rubric}
\label{app:sensitivity_annotation}

The unit of annotation is a candidate's edit relative to its round's
incumbent skill, not the benchmark as a whole. Each record is identified by
an \texttt{observation\_id} and linked to a unified skill diff. The diff-only
annotation workspace exposes \texttt{observation\_id},
\texttt{candidate\_diff}, and \texttt{diff\_file}; numerical screening gains
and candidate metadata are joined separately by the scorer. The annotator
is instructed not to use or infer benchmark, dataset, environment, or sample
identity, and to ignore intent names, acceptance decisions, optimizer
summaries, tool counts, and the outer interaction format.

A process annotation requires an added or strengthened ordered prerequisite,
state handoff, or conditional dependency in the solution procedure.
Generic advice to ``reason step by step'' is insufficient. An interference
annotation requires added guidance that is visibly duplicated, conflicting,
irrelevant to the surrounding skill, or disproportionately overlong.
A negative screening gain is not itself evidence of interference.
The same four-level strength scale is applied independently to both labels,
as detailed in \Cref{tab:sensitivity_rubric}.

\begin{table}[t]
\centering
\caption{Candidate-diff annotation rubric. Strength describes the visible
edit, not its measured benefit or the intrinsic complexity of the task.}
\label{tab:sensitivity_rubric}
\small
\begin{tabular}{@{}p{0.07\linewidth}p{0.42\linewidth}p{0.43\linewidth}@{}}
\toprule
\textbf{Score} & \textbf{Process structure} & \textbf{Interference} \\
\midrule
0 & No grounded addition or strengthening of a dependency.
  & No grounded addition of interfering guidance. \\
1 & Weak or local prerequisite, state handoff, or conditional dependency.
  & Weak or local duplication, conflict, irrelevance, or excessive detail. \\
2 & Clear dependency in the revised solution procedure.
  & Clear interfering guidance in the revision. \\
3 & Dependency structure is a dominant feature of the edit.
  & Interfering guidance is a dominant feature of the edit. \\
\bottomrule
\end{tabular}
\end{table}

The permitted interference modes are \texttt{duplicated},
\texttt{conflicting}, \texttt{irrelevant}, and \texttt{overlong}; more than
one may apply. Removing redundant text does not automatically receive a
positive interference label: the instruction concerns what the candidate
adds. If the diff is missing or supplies no grounded evidence, the prompt
requests zero strengths and an empty mode list.

The following illustrates the output schema, not a scored experimental
observation:
\begin{quote}
\small
\begin{verbatim}
{
  "trajectory_observations": [
    {
      "evidence_id": "<observation_id copied verbatim>",
      "evidence": "<specific evidence from the skill diff>",
      "process_structure_strength": 0,
      "interference_strength": 0,
      "interference_modes": []
    }
  ]
}
\end{verbatim}
\end{quote}

There must be exactly one observation per source candidate, with a unique
matching identifier and a nonempty evidence string. Both strengths must be
integers from 0 to 3; nonempty interference modes require a positive
interference strength. The annotator does not output PSS, RSS, confidence,
or a representation choice. The validated annotation runner allows up to
three generation attempts when output fails JSON or schema validation,
rather than accepting malformed output or heuristically repairing it.

\subsection{Eligibility, regression, and round bootstrap}
\label{app:sensitivity_estimation}

Annotations are joined to recorded candidate outcomes by identifier.
A candidate enters the regression if and only if it changes the incumbent,
passes patch validation, and has a recorded \texttt{screen\_net\_gain}.
Zero-gain and negative-gain candidates remain eligible, including candidates
rejected by the update gate. Unchanged or invalid candidates and candidates
without a screening outcome are excluded. The estimator uses screening
outcomes, not pooled confirmation gains or final test scores.

For \Cref{eq:normalized-edit-size}, the implementation compares the two
texts after \texttt{splitlines()} using
\texttt{difflib.\allowbreak SequenceMatcher} with \texttt{autojunk=False}. Each
non-equal block contributes the larger of its before and after line counts.
The denominator is the larger total line count, with a minimum of one for
the empty-text case.

For each eligible candidate, the numerical features are
\[
    (c_{t,j},\,p_{t,j},\,p_{t,j}^{2},\,r_{t,j},\,r_{t,j}^{2}).
\]
Each column is centered and divided by its population standard deviation;
a zero standard deviation is replaced by one. The design matrix also
includes an intercept and one-hot indicators for round and candidate
profile, omitting one reference level for each categorical variable.
There is no process--interference interaction term. With \(X\) denoting
this design and \(y\) the recorded screening gains, the fitted coefficients
are
\[
    \widehat{\beta}
    = (X^\top X+\Lambda)^{-1}X^\top y,
    \qquad \Lambda=\operatorname{diag}(0,1,\ldots,1).
\]
Thus the ridge penalty is one for every non-intercept coefficient,
including categorical controls. The implementation solves the linear
system directly rather than forming the inverse.

To evaluate a contrast, the scorer sets all rows' process and interference
strengths to the probe values, recomputes their squares, and retains each
row's edit size and categorical controls. It applies the fitted
standardization, averages the predictions, and computes the extrema in
\Cref{eq:trajectory-contrasts}. The full-sample fit yields
\(\Delta_{\mathrm{P}}\) and \(\Delta_{\mathrm{R}}\); the bootstrap yields
the positive-contrast proportions.

\begin{algorithm}[t]
\caption{Candidate-level PSS/RSS estimation}
\label{alg:sensitivity_estimator}
\begin{algorithmic}[1]
\small
\Input Joined eligible candidate rows \(\mathcal{D}\);
    bootstrap count \(N=2000\), random seed 42
\Output PSS \(\psi\), RSS \(\rho\), eligible count \(\nu\)
\State Let \(\mathcal{R}\) be the distinct rounds in \(\mathcal{D}\);
    set \(\nu=|\mathcal{D}|\)
\State Fit the full-sample ridge model and compute both contrasts
\For{\(n=1,\ldots,N\)}
    \State Draw \(|\mathcal{R}|\) rounds independently with replacement
        from \(\mathcal{R}\)
    \State Copy all eligible rows from each drawn round into
        \(\mathcal{D}^{(n)}\)
    \State Assign each drawn block a new round label, including repeated
        draws of the same source round
    \State Rebuild categorical levels and numerical standardization
        on \(\mathcal{D}^{(n)}\)
    \State Refit ridge and compute
        \(\Delta_{\mathrm{P}}^{(n)},\Delta_{\mathrm{R}}^{(n)}\)
\EndFor
\State Compute \(\psi,\rho\) using \Cref{eq:trajectory-sensitivity}
\State \Return \(\psi,\rho,\nu\)
\end{algorithmic}
\end{algorithm}

Sampling uses NumPy's \texttt{default\_rng(42)} over sorted round labels.
All eligible candidates from a drawn round stay together; candidates are
not independently resampled. The validated scoring runner checks that the
full-sample fit and all 2,000 bootstrap fits succeed, rather than silently
reporting a proportion over fewer fits. These resamples reuse existing round
blocks and are not 2,000 independent experiments. PSS/RSS summarize fitted
effect-sign stability, not the probability that one representation outperforms
another.

\section{Shared-Backbone Example}
\label{app:shared_backbone_example}

To isolate organization from content, we use the same ALFWorld backbone
for all three illustrations below. The source is the frozen seed-42 Phase~I
artifact shared by the ALFWorld ablation branches. Its \emph{Heat \& Place}
procedure reads:
\begin{quote}
Find X \(\rightarrow\) take X \(\rightarrow\) go to microwave
\(\rightarrow\) heat X \(\rightarrow\) go to Y \(\rightarrow\) put X.
\end{quote}
The same artifact instructs the agent to open closed containers before
judging them empty, transform an object before placing it, track progress,
avoid repeated-action loops, and choose only admissible actions.

\Cref{tab:shared_backbone_structures} re-expresses this source content
schematically. Unit names and dependency edges are introduced for
exposition; these are not verbatim exports of three learned checkpoints.
They illustrate representation semantics, not additional evaluation runs.
All three retain the same action-format and admissibility constraints.

\begin{table}[t]
\centering
\caption{Three schematic organizations of the same archived ALFWorld
guidance. The content is held fixed to make structural differences explicit.}
\label{tab:shared_backbone_structures}
\small
\begin{tabular}{@{}p{0.12\linewidth}p{0.80\linewidth}@{}}
\toprule
\textbf{Form} & \textbf{Organization of shared guidance} \\
\midrule
Flat &
Independently retrievable units for locating, acquiring, transforming,
and delivering the object, alongside exploration and progress/loop
guidance. No dependency graph schedules retrieval; a retrieved unit can
still state a local prerequisite such as ``transform before placing.'' \\
\addlinespace
Graph &
Explicit dependencies connect locate \(\rightarrow\) acquire
\(\rightarrow\) transform \(\rightarrow\) deliver.
Container inspection refines locating, while state and progress checks
provide guidance at the relevant nodes. Edges make prerequisite order
explicit. \\
\addlinespace
Hybrid &
The locate--acquire--transform--deliver chain forms the ordered core.
Container exploration, progress tracking, and loop recovery remain
auxiliary modules that can be retrieved when relevant, rather than
expanding every core step with all guidance. \\
\bottomrule
\end{tabular}
\end{table}

\paragraph{Observed execution evidence.}
\Cref{tab:shared_backbone_trace} shows one actual training episode from
the archived \texttt{round\_00\_train} batch, with source identifier
\texttt{train:0028}. Its goal is ``heat some egg and put it in countertop.''
Actions and zero-based step indices are retained verbatim; observations
are condensed for readability. The episode contains 12 actions and ends
with the environment marking the task complete. It was collected with
the textual Phase~I skill, not by executing each of the three schematic
representations.

\begin{table}[t]
\centering
\caption{Archived training execution illustrating state dependencies in
the shared backbone. This is one observed episode, not a three-way
representation comparison.}
\label{tab:shared_backbone_trace}
\small
\begin{tabular}{@{}p{0.06\linewidth}p{0.43\linewidth}p{0.43\linewidth}@{}}
\toprule
\textbf{Step} & \textbf{Action} & \textbf{Observed result (condensed)} \\
\midrule
0 & \texttt{go to fridge 1} & The fridge is closed. \\
1 & \texttt{open fridge 1} & Several objects are visible, but no egg. \\
2 & \texttt{go to microwave 1} & The microwave is closed. \\
3 & \texttt{open microwave 1} & The microwave is empty. \\
4 & \texttt{go to fridge 1} & The same fridge contents are observed again. \\
5 & \texttt{go to countertop 1} & An egg is visible among other objects. \\
6 & \texttt{take egg 1 from countertop 1} & The agent picks up the egg. \\
7 & \texttt{go to microwave 1} & The microwave is open. \\
8 & \texttt{heat egg 1 with microwave 1} & The environment confirms heating. \\
9 & \texttt{close microwave 1} & The microwave is closed. \\
10 & \texttt{go to countertop 1} & The agent reaches the destination. \\
11 & \texttt{move egg 1 to countertop 1} & The egg is placed; \texttt{done=true}. \\
\bottomrule
\end{tabular}
\end{table}

\paragraph{Reading the trace through the three representations.}
The handoff from acquisition at step~6 to heating at step~8, and then
delivery at step~11, makes the relevant state dependency concrete.
In the Flat illustration, the transformation unit supplies guidance after
acquisition without a graph-defined retrieval order. In Graph, explicit
edges connect those stages. In Hybrid, the same chain belongs to the core,
while container exploration or loop-recovery guidance can be supplied
separately. These describe how the evidence can be organized, not measured
retrieval choices for this episode. The return to the already inspected
fridge at step~4 is preserved in the trace: successful completion does not
imply perfect adherence to every instruction or optimal action efficiency.

\FloatBarrier
\section{Recorded Resource Usage}
\label{app:recorded_resource_usage}
\definecolor{tokenincrease}{HTML}{800020}

\Cref{tab:recorded_usage_gpt41,tab:recorded_usage_nano,tab:recorded_usage_gpt54,tab:recorded_usage_gpt56sol}
compare \skillopt{} and \ourmethod{} in terms of recorded token usage,
recorded calls, and final-test hard accuracy across four models and six
benchmarks.

\paragraph{Token calculation.}
Let \(\mathcal{C}\) denote the calls with recorded token usage in a run,
and let \(t_c^{\mathrm{in}}\) and \(t_c^{\mathrm{out}}\) be their reported
input and output token counts. We compute
\[
T_{\mathrm{in}}=\sum_{c\in\mathcal{C}}t_c^{\mathrm{in}},
\qquad
T_{\mathrm{out}}=\sum_{c\in\mathcal{C}}t_c^{\mathrm{out}},
\qquad
T_{\mathrm{total}}=T_{\mathrm{in}}+T_{\mathrm{out}}.
\]
Counts are summed within each run, not divided by the number of test
examples or averaged across runs. Each token column is divided by \(10^6\)
and rounded independently to three decimal places; totals are computed
before rounding. Missing token usage is not estimated by re-tokenizing
prompts or responses.
The column immediately following total tokens reports the relative change
\(\Delta_T=(T_{\mathrm{total}}^{\mathrm{SkillSpec}}/
T_{\mathrm{total}}^{\mathrm{SkillOpt}}-1)\times100\%\),
computed from unrounded counters and rounded to one decimal place.
Negative values (green) indicate fewer recorded tokens and positive values
(burgundy) indicate more; \skillopt{} is the reference and is marked with a dash.

\paragraph{Comparison.}
Within the reported counters, \ourmethod{} has fewer total tokens in
20 of the 24 model--benchmark pairs and fewer recorded calls in all 24.
Its final-test accuracy is higher in 23 of the 24 pairs.
These are descriptive comparisons, not end-to-end cost savings:
\skillopt{} counts may include built-in evaluations, whereas
\ourmethod{} counts cover only recorded Phase~II usage, excluding
Phase~I, representation estimation/selection, and unreported CLI tokens.
Recorded calls may include CLI invocations without token counts and thus
need not equal \(\lvert\mathcal{C}\rvert\) or complete provider API requests.

\begin{table}[htbp]
\centering
\caption{Recorded resource usage and test accuracy for GPT--4.1.
SkillSpec usage is partial Phase~II usage only; the accounting scopes are
not aligned for an end-to-end cost comparison.}
\label{tab:recorded_usage_gpt41}
\small
\setlength{\tabcolsep}{4pt}
\begin{tabular}{@{}llrrrrrr@{}}
\toprule
\textbf{Benchmark} & \textbf{Method} &
\shortstack{\textbf{Input}\\\textbf{(M)}} &
\shortstack{\textbf{Output}\\\textbf{(M)}} &
\shortstack{\textbf{Total}\\\textbf{(M)}} &
\shortstack{\textbf{Change}\\\textbf{(\%)}} &
\shortstack{\textbf{Recorded}\\\textbf{calls}} &
\shortstack{\textbf{Test}\\\textbf{(\%)}} \\
\midrule
SearchQA & \skillopt{} & 49.269 & 0.293 & 49.562 & \na & 15,948 & 79.93 \\
SearchQA & \ourmethod{} & 25.882 & 0.503 & 26.386 & \textcolor{deltapos}{\(-46.8\%\)} & 11,519 & 82.43 \\
\addlinespace
SpreadsheetBench & \skillopt{} & 61.921 & 4.329 & 66.251 & \na & 5,063 & 46.43 \\
SpreadsheetBench & \ourmethod{} & 68.802 & 3.274 & 72.075 & \textcolor{tokenincrease}{\(+8.8\%\)} & 3,941 & 50.36 \\
\addlinespace
OfficeQA & \skillopt{} & 21.136 & 0.232 & 21.368 & \na & 2,420 & 11.05 \\
OfficeQA & \ourmethod{} & 26.025 & 0.332 & 26.357 & \textcolor{tokenincrease}{\(+23.3\%\)} & 2,417 & 27.91 \\
\addlinespace
DocVQA & \skillopt{} & 4.184 & 0.053 & 4.237 & \na & 2,350 & 73.26 \\
DocVQA & \ourmethod{} & 2.368 & 0.042 & 2.410 & \textcolor{deltapos}{\(-43.1\%\)} & 1,432 & 84.49 \\
\addlinespace
LiveMath & \skillopt{} & 0.854 & 0.480 & 1.335 & \na & 585 & 29.03 \\
LiveMath & \ourmethod{} & 2.002 & 0.514 & 2.516 & \textcolor{tokenincrease}{\(+88.5\%\)} & 393 & 41.94 \\
\addlinespace
ALFWorld & \skillopt{} & 30.389 & 6.602 & 36.991 & \na & 28,543 & 52.99 \\
ALFWorld & \ourmethod{} & 23.300 & 2.628 & 25.928 & \textcolor{deltapos}{\(-29.9\%\)} & 11,966 & 79.10 \\
\bottomrule
\end{tabular}
\end{table}

\begin{table}[htbp]
\centering
\caption{Recorded resource usage and test accuracy for GPT--5.4 Nano.
SkillSpec usage is partial Phase~II usage only; the accounting scopes are
not aligned for an end-to-end cost comparison.}
\label{tab:recorded_usage_nano}
\small
\setlength{\tabcolsep}{4pt}
\begin{tabular}{@{}llrrrrrr@{}}
\toprule
\textbf{Benchmark} & \textbf{Method} &
\shortstack{\textbf{Input}\\\textbf{(M)}} &
\shortstack{\textbf{Output}\\\textbf{(M)}} &
\shortstack{\textbf{Total}\\\textbf{(M)}} &
\shortstack{\textbf{Change}\\\textbf{(\%)}} &
\shortstack{\textbf{Recorded}\\\textbf{calls}} &
\shortstack{\textbf{Test}\\\textbf{(\%)}} \\
\midrule
SearchQA & \skillopt{} & 43.717 & 0.468 & 44.185 & \na & 16,339 & 57.43 \\
SearchQA & \ourmethod{} & 48.937 & 0.130 & 49.067 & \textcolor{tokenincrease}{\(+11.0\%\)} & 11,218 & 71.79 \\
\addlinespace
SpreadsheetBench & \skillopt{} & 117.841 & 7.458 & 125.299 & \na & 5,739 & 39.64 \\
SpreadsheetBench & \ourmethod{} & 95.617 & 5.098 & 100.715 & \textcolor{deltapos}{\(-19.6\%\)} & 4,243 & 45.71 \\
\addlinespace
OfficeQA & \skillopt{} & 21.265 & 0.294 & 21.559 & \na & 2,558 & 5.81 \\
OfficeQA & \ourmethod{} & 9.846 & 0.166 & 10.012 & \textcolor{deltapos}{\(-53.6\%\)} & 1,370 & 7.56 \\
\addlinespace
DocVQA & \skillopt{} & 5.236 & 0.101 & 5.337 & \na & 2,692 & 65.51 \\
DocVQA & \ourmethod{} & 3.215 & 0.110 & 3.325 & \textcolor{deltapos}{\(-37.7\%\)} & 1,526 & 66.31 \\
\addlinespace
LiveMath & \skillopt{} & 1.701 & 0.100 & 1.801 & \na & 654 & 50.81 \\
LiveMath & \ourmethod{} & 1.594 & 0.066 & 1.660 & \textcolor{deltapos}{\(-7.8\%\)} & 417 & 52.42 \\
\addlinespace
ALFWorld & \skillopt{} & 53.174 & 4.135 & 57.309 & \na & 33,051 & 33.58 \\
ALFWorld & \ourmethod{} & 19.984 & 1.417 & 21.400 & \textcolor{deltapos}{\(-62.7\%\)} & 13,845 & 48.51 \\
\bottomrule
\end{tabular}
\end{table}

\begin{table}[htbp]
\centering
\caption{Recorded resource usage and test accuracy for GPT--5.4.
SkillSpec usage is partial Phase~II usage only; the accounting scopes are
not aligned for an end-to-end cost comparison.}
\label{tab:recorded_usage_gpt54}
\small
\setlength{\tabcolsep}{4pt}
\begin{tabular}{@{}llrrrrrr@{}}
\toprule
\textbf{Benchmark} & \textbf{Method} &
\shortstack{\textbf{Input}\\\textbf{(M)}} &
\shortstack{\textbf{Output}\\\textbf{(M)}} &
\shortstack{\textbf{Total}\\\textbf{(M)}} &
\shortstack{\textbf{Change}\\\textbf{(\%)}} &
\shortstack{\textbf{Recorded}\\\textbf{calls}} &
\shortstack{\textbf{Test}\\\textbf{(\%)}} \\
\midrule
SearchQA & \skillopt{} & 57.311 & 3.468 & 60.779 & \na & 16,074 & 84.00 \\
SearchQA & \ourmethod{} & 49.474 & 2.758 & 52.232 & \textcolor{deltapos}{\(-14.1\%\)} & 11,060 & 86.36 \\
\addlinespace
SpreadsheetBench & \skillopt{} & 10.552 & 9.144 & 19.696 & \na & 2,282 & 71.43 \\
SpreadsheetBench & \ourmethod{} & 7.756 & 6.633 & 14.389 & \textcolor{deltapos}{\(-26.9\%\)} & 1,511 & 76.79 \\
\addlinespace
OfficeQA & \skillopt{} & 25.875 & 3.468 & 29.342 & \na & 2,620 & 59.88 \\
OfficeQA & \ourmethod{} & 17.448 & 2.540 & 19.987 & \textcolor{deltapos}{\(-31.9\%\)} & 1,979 & 66.28 \\
\addlinespace
DocVQA & \skillopt{} & 4.877 & 0.441 & 5.317 & \na & 2,568 & 90.37 \\
DocVQA & \ourmethod{} & 2.023 & 0.401 & 2.424 & \textcolor{deltapos}{\(-54.4\%\)} & 1,450 & 88.77 \\
\addlinespace
LiveMath & \skillopt{} & 2.137 & 3.685 & 5.823 & \na & 782 & 55.65 \\
LiveMath & \ourmethod{} & 1.083 & 2.235 & 3.319 & \textcolor{deltapos}{\(-43.0\%\)} & 417 & 58.87 \\
\addlinespace
ALFWorld & \skillopt{} & 24.464 & 4.832 & 29.296 & \na & 14,327 & 89.55 \\
ALFWorld & \ourmethod{} & 10.397 & 1.905 & 12.302 & \textcolor{deltapos}{\(-58.0\%\)} & 5,796 & 92.54 \\
\bottomrule
\end{tabular}
\end{table}

\begin{table}[htbp]
\centering
\caption{Recorded resource usage and test accuracy for GPT--5.6 Sol.
SkillSpec usage is partial Phase~II usage only; the accounting scopes are
not aligned for an end-to-end cost comparison.}
\label{tab:recorded_usage_gpt56sol}
\small
\setlength{\tabcolsep}{4pt}
\begin{tabular}{@{}llrrrrrr@{}}
\toprule
\textbf{Benchmark} & \textbf{Method} &
\shortstack{\textbf{Input}\\\textbf{(M)}} &
\shortstack{\textbf{Output}\\\textbf{(M)}} &
\shortstack{\textbf{Total}\\\textbf{(M)}} &
\shortstack{\textbf{Change}\\\textbf{(\%)}} &
\shortstack{\textbf{Recorded}\\\textbf{calls}} &
\shortstack{\textbf{Test}\\\textbf{(\%)}} \\
\midrule
SearchQA & \skillopt{} & 40.721 & 0.691 & 41.412 & \na & 15,219 & 85.57 \\
SearchQA & \ourmethod{} & 23.513 & 0.287 & 23.800 & \textcolor{deltapos}{\(-42.5\%\)} & 11,186 & 86.29 \\
\addlinespace
SpreadsheetBench & \skillopt{} & 12.834 & 5.637 & 18.471 & \na & 2,512 & 82.14 \\
SpreadsheetBench & \ourmethod{} & 8.943 & 3.008 & 11.951 & \textcolor{deltapos}{\(-35.3\%\)} & 1,520 & 82.50 \\
\addlinespace
OfficeQA & \skillopt{} & 40.449 & 1.933 & 42.382 & \na & 4,208 & 54.65 \\
OfficeQA & \ourmethod{} & 15.564 & 0.756 & 16.320 & \textcolor{deltapos}{\(-61.5\%\)} & 2,097 & 65.70 \\
\addlinespace
DocVQA & \skillopt{} & 10.070 & 0.102 & 10.172 & \na & 2,552 & 90.37 \\
DocVQA & \ourmethod{} & 5.091 & 0.028 & 5.119 & \textcolor{deltapos}{\(-49.7\%\)} & 1,398 & 91.98 \\
\addlinespace
LiveMath & \skillopt{} & 1.518 & 0.766 & 2.284 & \na & 778 & 67.74 \\
LiveMath & \ourmethod{} & 0.580 & 0.295 & 0.875 & \textcolor{deltapos}{\(-61.7\%\)} & 387 & 76.61 \\
\addlinespace
ALFWorld & \skillopt{} & 17.015 & 1.372 & 18.387 & \na & 11,612 & 88.06 \\
ALFWorld & \ourmethod{} & 7.941 & 0.489 & 8.430 & \textcolor{deltapos}{\(-54.2\%\)} & 5,416 & 90.30 \\
\bottomrule
\end{tabular}
\end{table}

\end{document}